%% file: main.tex
\documentclass[sigconf]{acmart}

\AtBeginDocument{%
  \providecommand\BibTeX{{%
    \normalfont B\kern-0.5em{\scshape i\kern-0.25em b}\kern-0.8em\TeX}}}

\usepackage{microtype}
\usepackage{graphicx}
\usepackage{caption}
\usepackage{subcaption}
\usepackage{booktabs} 
\usepackage{tabularx}
\usepackage{multicol}
\usepackage{empheq}
\usepackage{amsmath}
\usepackage{todonotes}
\usepackage{xspace} 
\usepackage{multirow}
\usepackage{dirtytalk}
\usepackage{algorithmic}
\usepackage{tikz}
\usepackage[
  separate-uncertainty = true,
  multi-part-units = repeat
]{siunitx}
\usepackage{adjustbox}

\usepackage{fancyhdr}
\setcopyright{none}
\renewcommand\footnotetextcopyrightpermission[1]{}
\newcommand{\approach}{{\sc CamTuner}\xspace}

\newcommand{\etc}{\emph{etc.}\xspace}

\newcommand{\ie}{\emph{i.e.,}\xspace}
\newcommand{\eg}{\emph{e.g.,}\xspace}
\newcommand{\etal}{\emph{et al.}\xspace}
\renewcommand{\comment}[1]{{\em \color{blue} {#1}}}

\newcommand{\cut}[1]{}

\usepackage{hyperref}

\newcommand{\figref}[1]{Figure \ref{#1}}
\newcommand{\secref}[1]{Section \ref{#1}}
\newcommand{\tabref}[1]{Table \ref{#1}}

\begin{document}

\title{Enhancing Video Analytics Accuracy via Real-time Automated Camera Parameter Tuning}
\author{Sibendu Paul}
\affiliation{%
  \institution{Purdue University}
  \city{West Lafayette}
  \country{USA}
}
\email{paul90@purdue.edu}

\authornote{To appear in \textbf{ACM Sensys 2022}}

\author{Kunal Rao}
\affiliation{%
  \institution{NEC Laboratories America, Inc.}
   \city{New Jersey}
  \country{USA}
}
\email{kunal@nec-labs.com}

\author{Giuseppe Coviello}
\affiliation{%
  \institution{NEC Laboratories America, Inc.}
   \city{New Jersey}
  \country{USA}
}
\email{giuseppe.coviello@nec-labs.com}

\author{Murugan Sankaradas}
\affiliation{%
  \institution{NEC Laboratories America, Inc.}
   \city{New Jersey}
  \country{USA}
}
\email{murugs@nec-labs.com}

\author{Oliver Po}
\affiliation{%
  \institution{NEC Laboratories America, Inc.}
   \city{San Jose}
  \country{USA}
}
\email{oliver@nec-labs.com}

\author{Y. Charlie Hu}
\affiliation{%
  \institution{Purdue University}
  \city{West Lafayette}
  \country{USA}
}
\email{ychu@purdue.edu}

\author{Srimat Chakradhar}
\affiliation{%
  \institution{NEC Laboratories America, Inc.}
   \city{New Jersey}
  \country{USA}
}
\email{chak@nec-labs.com}


\vskip 0.3in
\renewcommand{\shortauthors}{Sibendu Paul, et al.}
\input{abstract}

\maketitle

\input{introduction}

\input{background}

\input{motivation}

\input{challenges}

\input{design}

\input{impl}

\input{eval}

\input{related}

\input{future}

\input{conclusion}


\bibliographystyle{abbrv}
\bibliography{egbib}



\end{document}

%% file: abstract.tex
\begin{abstract}

In Video Analytics Pipelines (VAP), Analytics Units (AUs) such as
object detection
and face recognition running on
remote servers critically rely on surveillance cameras to capture
high-quality video streams in order to achieve high accuracy.
Modern IP cameras come with a large number of camera parameters
that directly affect the quality of the video
stream capture.  While a few of such parameters, 
\eg exposure, focus, white balance are automatically adjusted  by the camera internally, the remaining ones are not.
We denote such camera parameters as non-automated (NAUTO) parameters.
In this paper, we first show that environmental
condition changes can have significant adverse effect on the accuracy
of insights from the AUs, but such adverse impact can potentially
be mitigated by dynamically adjusting NAUTO camera parameters in response to
changes in environmental conditions.  
We then present \approach, to our knowledge, 
the first framework that
dynamically adapts NAUTO camera parameters to
optimize the accuracy of AUs in a VAP in response to adverse changes in
environmental conditions.  \approach is based on SARSA reinforcement
learning and it incorporates two novel components: a light-weight
analytics quality estimator and a virtual camera that drastically 
speed up offline RL training.
%
Our controlled experiments and real-world VAP deployment
show that compared to a VAP using the default camera setting,
\approach enhances VAP accuracy by detecting 
15.9\%
additional persons 
and
2.6\%--4.2\% additional cars
(without any false positives)
in a large enterprise parking lot
and 9.7\% additional cars in a 5G smart traffic intersection scenario,
which enables a new usecase of accurate and reliable automatic vehicle collision prediction (AVCP). 
\approach opens doors for new ways to significantly enhance video
analytics accuracy beyond incremental improvements 
from refining deep-learning models.

\end{abstract}

%% file: introduction.tex
\section{Introduction}
\label{section:intro}

Significant progress in machine learning and computer vision techniques
for analyzing video streams \cite{imagenet}, along with the explosive growth in Internet
of Things (IoT), edge computing, and high-bandwidth access networks
such as 5G~\cite{QUALCOMM-5G, CNET-5G}, have led to the wide adoption
of video analytics systems. Such systems deploy cameras throughout the
world to support diverse applications in entertainment, health-care,
retail, automotive, transportation, home automation, safety, and
security market segments.
The global video analytics market is estimated to grow from \$5
billion in 2020 to \$21 billion by 2027, at a CAGR of
22.70\%~\cite{allied-market-research}.

\begin{figure}[tb]
    \centering
    \includegraphics[width=0.95\linewidth]{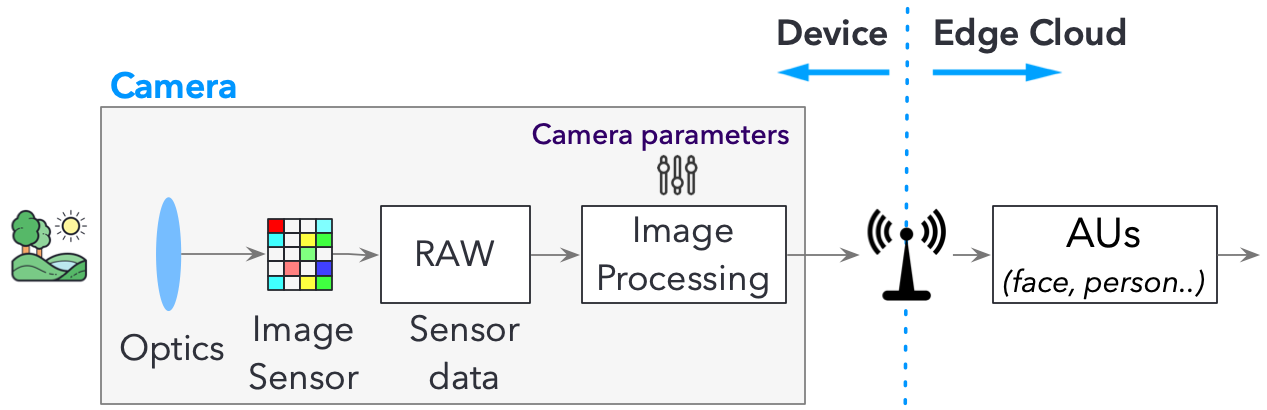}
    \caption{Video analytics pipeline.}
    \label{fig:pipeline}
\vspace{-0.2in}
\end{figure}

A typical video analytics system consists of a video analytics
pipeline (VAP) that starts with one or more surveillance cameras capturing live
feed of the target environment. These live feeds are sent over a 5G
network to servers at the edge of the 5G network 
where one or more analytics units (AUs) such as object detection, face
detection, and face recognition use deep learning models to mine
valuable information in the live video streams, as shown in
Figure~\ref{fig:pipeline}.  These AUs critically rely on the 
cameras to capture high-quality real-time video streams in order to
achieve high accuracy.

Modern IP cameras come with and expose a large number of camera parameters 
that direcly affect the quality of the video
stream capture.  While a few of such parameters, \eg exposure, focus, white balance are automatically adjusted  by the camera internally, the
remaining camera parameters are not.
We denote such camera parameters as {\em non-automated (NAUTO) parameters}.

In this paper, we first show that as the environmental conditions
around the cameras change, the quality of video frames captured by the
cameras also changes, and this can adversely affect the accuracy of
insights derived by the analytics units. 
In our experiments, we kept all {automatic} parameter setting features
turned on and thus our experiments show that those automatic settings
are not enough to adapt to different environments for better analytics accuracy

Next, we experimentally show that by (manually) dynamically adjusting
a prominent set of NAUTO camera parameters, in particular, four image appearance
parameters including brightness, contrast, color-saturation (also
known as colorfulness), and {sharpness},
which are available in both PTZ and non-PTZ cameras,
it is possible to mitigate the potential loss in accuracy due to
adverse environmental changes. We chose these NAUTO parameters in our study
because they not only directly affect image qualities and hence
AU accuracy but also are challenging to tune due to their large ranges of values.

Since streaming video analytics systems operate around the clock
(24 hours a day, seven days a week), it is not practical for
humans to manually adjust tens of configurable camera parameters in
real-time in response to every environmental change. 
Therefore, we propose \approach, a system
that detects and dynamically adapts to the changes in environmental
conditions by automatically adjusting camera parameters in real-time to
improve AU accuracy. 
%
\approach uses online reinforcement learning
(RL)~\cite{introduction-to-rl} to continuously learn good camera
settings and update the camera parameters to enhance the accuracy of
the AUs in the VAP. In particular, \approach uses
SARSA~\cite{q-lambda}, which is faster to train and achieves slightly
better accuracy in our video stream processing context than other popular RL
approaches like Q-learning.

Although RL is a fairly standard technique, applying it to tuning
camera parameters in a real-time video analytics system poses two
unique challenges.

\textit{First}, implementing online RL requires knowing the
reward/penalty for every action taken during exploration and
exploitation. Since no ground truth for an AU task like face detection
is available during the online operation of a VAP, calculating the
reward/penalty due to an action taken by an RL agent is a key
challenge. To address this challenge, we propose an {\em AU-specific
  analytics quality estimator} that can accurately estimate the
accuracy of the AU. Our estimator is light-weight, and it can run on a
low-end PC or a simple IoT device to process video streams in
real-time.

\textit{Second}, bespoke online RL learning at each camera deployment
setup requires initial {\em RL training}, which can potentially take a long
time for two reasons: (1) capturing the environmental condition
changes such as the time-of-the-day effect can take a long time, and
(2) taking an action on the real camera (\ie changing the camera parameter
setting) by using the APIs provided by the camera vendor incurs a significant delay of about 200 ms. This limits the
speed of state transitions during RL exploration, and hence the training speed of RL, to
about 5 changes (actions) per second. To address these two sources of
RL training inefficiencies, we propose a novel concept called {\em virtual
  camera}. 
A virtual camera mimics (in software) the effect of changing parameters of a physical camera to capture a scene.
There are two key benefits of doing this: (1) we can complete an action
of ``camera setting change'' almost instantaneously; and (2) we can
digitally {\em augment} a single frame captured by the real camera to derive many new
synthetically transformed frames, {as if we had physically captured many different frames of the same scene by using a real camera 
at different environmental conditions (\ie time-of-day, lighting conditions, seasonal changes \etc)}. 
These two benefits allow
the RL agent to explore actions at a much faster rate than possible in using a
real camera. This drastically reduces the RL training time required to develop a good, initial RL model, which can then be further refined 
in a short period (adaptation phase) after camera deployment. 

Our paper makes the following contributions:
\begin{itemize}
    \item We show that environmental condition changes can have a
      significant negative impact on the accuracy of AUs in video
      analytics pipelines, but the negative impact can be mitigated by
      dynamically adjusting a set of NAUTO camera parameters.
    \item We develop, to our knowledge, the first system that
      automatically and adaptively learns and tunes the set of NAUTO camera parameters
      in response to unpredictable environmental condition changes to
      improve the accuracy of insights from video analytics pipelines.
    \item We present two novel techniques that make the RL-based
      camera-parameter-tuning design feasible: a light-weight
      AU-specific analytics quality estimator that enables online RL
      without requiring ground truth, and a virtual camera that
      enables fast initial RL model training.
\item We  show that \approach improves AU accuracy in 
controlled experiments and in real VAP
  deployment.  In particular, in a real world deployment where
  two cameras deployed side-by-side (one camera is managed by
  \approach, while the other is not) are monitoring a large enterprise
  parking lot, and the live video streams are carried over a 5G network,
the camera managed by \approach  detected
15.9\% (146)
  additional persons (in a 5-minute span) during
  evening hours, without any false positives.
The camera managed by \approach detected 
2.6\%--4.2\% 
(861--881) additional cars (in a 5-minute span) during morning and
evening hours, again without any false positives.
\item
Furthermore, by recording a real-world car accident scenario at a traffic intersection (at one of
our customer locations) and by using VC to emulate frame captures at different times of the
day, the VAP with \approach reliably detected 9.7\% (122) additional cars (across the frames in a 1.5-minute span), which dramatically improves the accuracy (and lead time) of collision prediction. 
\item 
We show that \approach incurs very low computation overhead and \approach 
can be easily incorporated into VAPs that are executing on low-end PC or IoT devices that are directly attached to the camera.
\end{itemize}

%% file: background.tex
\section{Background}
\label{section:background}

\figref{fig:pipeline} also shows the image signal processing (ISP) pipeline within a camera. 
Photons from the external world reach the image sensor through an
optical lens. The image sensor uses a Bayer
filter~\cite{bayer1976color} to create raw-image data, which is
further enhanced by a variety of image processing techniques such as
demosaicing, denoising, white balance, color-correction, sharpening
and image compression (JPEG/PNG or video compression using
H.264~\cite{x264}, VP9~\cite{vp9}, MJPEG, \etc) in the image-signal
processing (ISP) stage~\cite{ramanath2005color} before the camera
outputs an image or a video frame.

The camera capture forms the initial stage of the VAP, which may
include a wide variety of analytics tasks such as {face detection,
  face recognition, human pose estimation, license plate recognition}
\etc (see Figure~\ref{fig:pipeline}).

\cut{ and has found a wide range of application domains including
  entertainment, health-care, retail, automotive, transport, home
  automation, safety, and security.  }

\begin{table}[tp]
\caption{Parameters exposed by popular cameras. 
Parameters with ``*'' are auto-adjusted by the camera internally.
}
\vspace{-0.1in}
\label{tab:cameras}
{
\parbox{0.435\linewidth}{
\small
\begin{tabular}{|c|c|}
\hline
\multicolumn{2}{|c|}{Camera Setting Parameters}\\
\hline
\multirow{4}{*}{Image} & Brightness\\
\multirow{4}{*}{Appearance}  & sharpness\\
 & contrast \\
 & color level \\
 \hline
  \multirow{5}{*}{Exposure} & Exposure Control$^*$ \\
\multirow{5}{*}{Settings} & Max Exposure Time \\
& Exposure Zones$^*$ \\
& Max gain \\
& IR cut filter$^*$ \\
\hline
\multirow{4}{*}{Image} & Defog Effect \\
\multirow{4}{*}{Correction} & Noise Reduction \\
& Stabilizer \\
& Auto Focus Enabled$^*$ \\
\hline
White & Type$^*$\\
Balance & window$^*$ \\
\hline
\end{tabular}
}
\hfill
\parbox{0.435\linewidth}{
\small
\begin{tabular}{|c|c|}
\hline
\multicolumn{2}{|c|}{Video Stream Parameters}\\
\hline
& \\
\multirow{3}{*}{Image} & Resolution \\
\multirow{3}{*}{Appearance} & Compression \\
& Rotate image \\
& \\
\hline
\multirow{2}{*}{Encoder} & GOP length \\
\multirow{2}{*}{Settings} & H.264 profile \\
 & \\
\hline
 & \\
\multirow{3}{*}{Bitrate} & Type of Use \\
\multirow{3}{*}{Control} & Target Bitrate \\
& Priority \\
& \\
\hline
Video Stream & Max FPS \\
\hline
MJPEG & Max frame size \\
\hline
\end{tabular}
}
}
\end{table}

In this paper, we study video analytics applications that are based on
surveillance cameras. Such cameras are running 24X7 in contrast to
DSLR, point-and-shoot or mobile cameras that capture videos
on-demand. Popular IP video surveillance cameras are manufactured by
vendors such as AXIS~\cite{axis}, Cisco~\cite{ciscocamera}, and
Panasonic~\cite{panasonic}. 
These surveillance camera manufacturers have exposed many
camera parameters via REST APIs
which can be set by applications to control
the image generation process, which in turn affects the quality of
the produced image or video. 
The exposed parameters include those for changing the
amount of light that hits the sensor, the zoom level and
field-of-view (FoV) at the image-sensor stage, and those for changing
the color-saturation, brightness, contrast, sharpness, gamma,
acutance, \etc in the ISP stage.
Table~\ref{tab:cameras} lists the parameters exposed by a few popular surveillance cameras in the market today.
Remotely changing the camera setting via the exposed APIs,
however, incurs a significant delay, \eg about 200 ms on
Axis Q1615, Axis Q3515, Axis Q6128-E and Axis Q3505 MK II network camera.

While a few of these camera parameters, \eg exposure, focus, balance,
are automatically adjusted by the camera internally, the 
remaining camera parameters are not adjusted automatically.  We
denote such camera parameters as {\em non-automated (NAUTO) parameters}.

In this paper, we focus our study on the four image appearance camera
parameters, denoted as {\em I-A parameters} in the rest of the paper,
which are widely available in both PTZ and non-PTZ cameras:
{\em brightness, contrast, color-saturation} (also known as
colorfulness), and {\em sharpness}.
We chose the above four NAUTO camera parameters in our study in this
paper for two reasons: (1) they directly affect the quality of the image which is essential to AUs which typically extract insights,
\eg face recognition, from individual frames;
(2) These parameters are more challenging to tune due to the large range
(for example, between 1 and 100 for each of the parameters on Axis Q1615, Axis Q3515, Axis Q6128-E, Axis Q3505 MK II network camera \etc) compared to other NAUTO camera parameters which have either a few fixed settings or just a binary {ON}/{OFF} switch.


%% file: motivation.tex
\section{Motivation}
\label{section:motivation}

We motivate the need for dynamically adjusting NAUTO camera settings by
experimentally showing the impact of environmental changes on AU
accuracy despite all the auto-setting features are left on, and that
tuning a set of NAUTO camera settings can improve AU accuracy under the
same environmental conditions. 
%

\vspace{-0.05in}
\subsection{Impact of Environment Change on AU Accuracy}
\label{subsec:environ}

\begin{figure}[tb]
\begin{subfigure}[]{0.48\linewidth}
\centering
    \includegraphics[height=1.25 in]{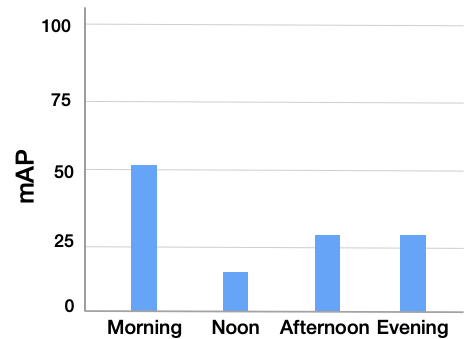}
    \caption{Face detection}
    \label{fig:daylong_face_det}
\end{subfigure}%
\begin{subfigure}[]{0.48\linewidth}
\centering
    \includegraphics[height=1.25 in]{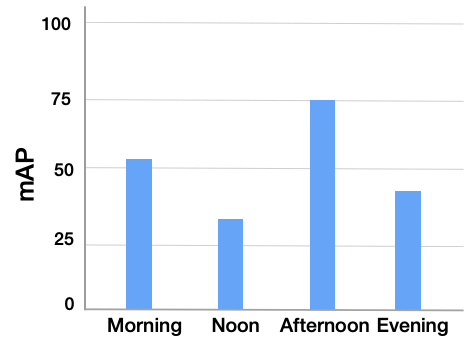}
    \caption{Person detection}
    \label{fig:daylong_person_det}
\end{subfigure}
    \vspace{-0.1in}
 \caption{AU accuracy variation in a day under the default camera setting.}
  \label{fig:face_person_env}
 \vspace{-0.1in}
\end{figure}

Environmental changes happen for at least three reasons.
First, such changes can be induced due to the change of
the Sun's movement throughout a day, \eg sunrise and sunset.
Second, they can be triggered by 
changes in weather conditions, \eg rain, fog, and snow.
Third, even for the same weather condition at exactly the same
time of the day, the videos captured by the cameras at different 
deployment sites (\eg parking lot, factory, shopping mall, and airport)
can have diverse content and ambient lighting conditions.

To illustrate the impact of environmental changes on image quality, and
consequently on the accuracy of AUs, we experimentally measure the
accuracy of two popular AUs (face detection and person detection)
throughout a 24-hour (one-day) period.  Since there are
no publicly available video datasets that capture the environmental
variations in a day or a week by using the same camera (at the same location), we use
several proprietary videos provided by our customers that 
were captured with the default camera setting -- in this paper,
{\em the default camera setting} refers to when all auto-setting features are turned on
and NAUTO parameters are set to the default values provided by the manufacturers.
These videos were captured outside airports and baseball stadiums by
stationary surveillance cameras, 
and we have labeled ground-truth information for several analytics tasks
 including face detection and person detection.

We use RetinaNet~\cite{deng2019retinaface} for face detection and
EfficientDet-v8~\cite{tan2020efficientdet} for person detection. 
We compute the mean Average Precision (mAP) by using pycocotools~\cite{pycoco}. \figref{fig:daylong_face_det} shows that the average mAP values for the
face detection AU during four different time periods of the day
(morning 8AM - 10AM, noon 12PM - 2PM, afternoon 3PM -
  5PM, and evening 6PM - 8PM), and with the default camera 
setting, can vary by up to 40\% as the day progresses (blue bars).
Similarly, \figref{fig:daylong_person_det} shows that the average mAP
values for the person detection AU (with the default camera parameter
setting) can vary by up to 38\% during the four time periods.
These results show that changes in environmental
conditions can adversely affect the quality of the frames retrieved from the
camera, and consequently adversely impact the accuracy of the insights that are
derived from the video data.

\vspace{-0.1in}
\subsection{Impact of Image Appearance Camera Settings on AU Accuracy}
\label{subsec:impactofsettings}

We experimentally show that adjusting the four image appearance (I-A)
(NAUTO) camera settings, \ie brightness, contrast, color-saturation (also
known as colorfulness), and {\em sharpness}, can help to mitigate the
adverse impact of environmental changes on AU accuracy.

Analyzing the impact of camera settings on video analytics
in general faces a significant challenge: it requires applying
different camera parameter settings to the same input scene and
measuring the resulting accuracy of insights from an AU.
The straightforward approach is to use multiple cameras with different
camera parameter settings to capture the same input scene. However,
this approach is impractical as there are thousands of different
combinations of even just the four camera parameters we consider.  To
overcome the challenge, we proceed with the following workaround
which uses a single real camera.


We use a real camera,
Axis Q3505 MK II Network camera, to capture (at 10 FPS) the same
real-world scene repeatedly under varying camera settings,
%
and 
compare the accuracy of object detection AU for "default" and several "modified"
settings -- in this paper, a {\em "modified" setting}
refers to modifying the four I-A camera settings while keeping
all other camera parameter values the same as the default setting.
In our scene, two people walk from the camera
towards two parked cars, and each of them then starts driving a separate car in
a loop within the parking lot, parks the car in the same parking
spot, and walks back towards the camera. The entire sequence of steps
takes around 2 minutes and we repeat these exact steps over and over
again for 26 different camera settings (including the "default"
setting). These experiments are conducted immediately one after
another in quick succession to minimize the effect of environmental
change.

\begin{figure}[tb]
\begin{subfigure}[]{0.48\linewidth}
\centering
    \includegraphics[width=1.54 in]{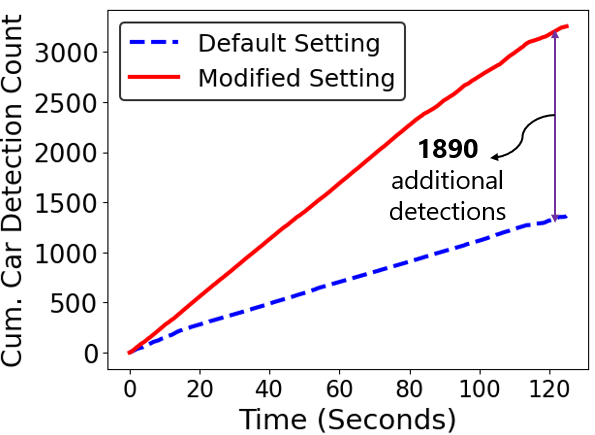}
    \caption{Car Detection}
    \label{fig:cardet_improv}
\end{subfigure}%
\begin{subfigure}[]{0.48\linewidth}
\centering
    \includegraphics[width=1.52 in]{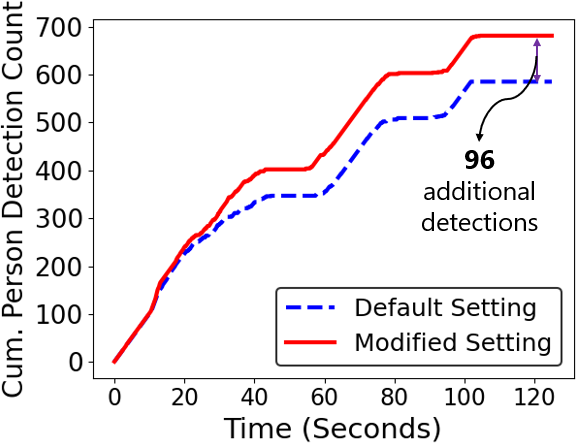}
    \caption{Person Detection}
    \label{fig:persondet_improv}
\end{subfigure}
\vspace{-0.1in}
 \caption{Impact of camera settings on object detection.}
  \label{fig:left_right_obj}
 \vspace{-0.15in}
\end{figure}


\begin{figure}[t]
    \centering
    \includegraphics[width=0.95\linewidth]{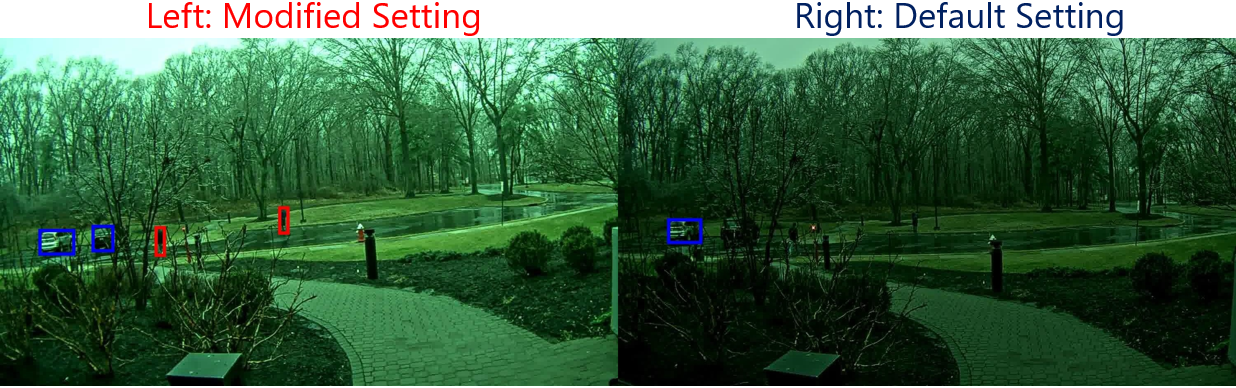}
    \vspace{-0.1in}
    \caption{Same scene with different camera settings.}
    \label{fig:left_right_cap}
\vspace{-0.1in}
\end{figure}
 
We consider  object detection AU in this experiment, which
detects cars and persons in the scene. Specifically, we use
Efficientdet~\cite{tan2020efficientdet} object detector.
An illustration of the
scene is shown in \figref{fig:left_right_cap} with two side-by-side
frames, where the right one is with the "default" camera setting and the
left one is with a "modified" camera setting. Here, the AU can accurately
detect two persons and two cars on the frame with the modified camera
setting while from the camera capture under the default setting, the AU can
only detect one car. We observe that the accuracy of the AU
varies across different camera settings and
\figref{fig:left_right_obj} shows the \textit{cumulative number} of
true-positive car and person detection counts~\footnote{An 
  object detection is true-positive if the detector correctly predicts the
  object label and the IoU between the detected and ground-truth
  bounding box is more than 0.7.} for "Default Setting" and for 
"Modified Setting", which shows the highest accuracy among the 25
different camera settings. We see that "Modified Setting"
correctly detected 1890 additional cars and 96 additional persons across
1300 frames compared to ``Default Setting''.

\begin{table}[tb]
\begin{center}
\caption{Best settings for different environment.}
\vspace{-0.1in}
\label{tab:diff-env-settings}
{
\small
\begin{tabular}{|c|c|}
    \hline
    Environment & Best Camera Setting\\
    (Time-of-day) & [brightness, contrast, color, sharpness]\\
    \hline
    Dawn & {[80, 75, 50, 75]}  \\
    \hline
   Morning & [30, 30, 50, 50]  \\
    \hline
     Evening & [90, 90, 50, 50]\\
    \hline
\end{tabular}
}
\end{center}
\vspace{-0.2in}
\end{table}

To understand if the modified camera setting that provides the
highest AU accuracy remains the same as the environment undergoes
changes, we repeat the above experiment for three different times of
the day, \ie \textit{dawn}, \textit{morning} and \textit{evening},
which have varying sunlight, while enacting the same scene for camera
capture.
%
~\tabref{tab:diff-env-settings} shows that the I-A camera
setting that provides the highest AU accuracy is not the same as the
manufacturer-provided default setting and it varies for different
times of the day, \ie different environmental conditions.

\cut{
\begin{figure}[tb]
\begin{subfigure}[]{0.5\linewidth}
\centering
    \includegraphics[width=1.6 in]{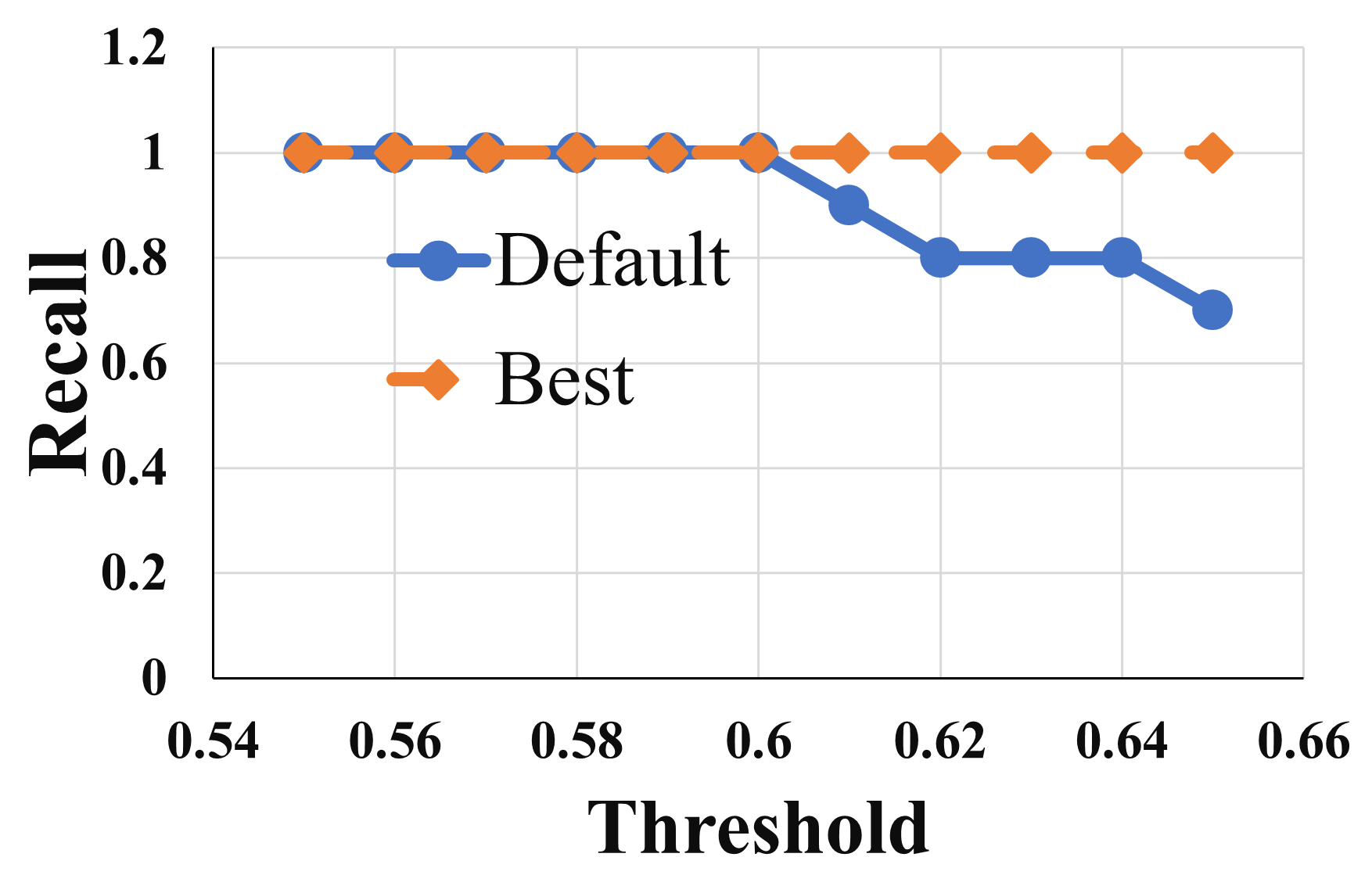}
    \caption{DAY}
    \label{fig:recall_day}
\end{subfigure}%
\begin{subfigure}[]{0.5\linewidth}
\centering
    \includegraphics[width=1.6 in]{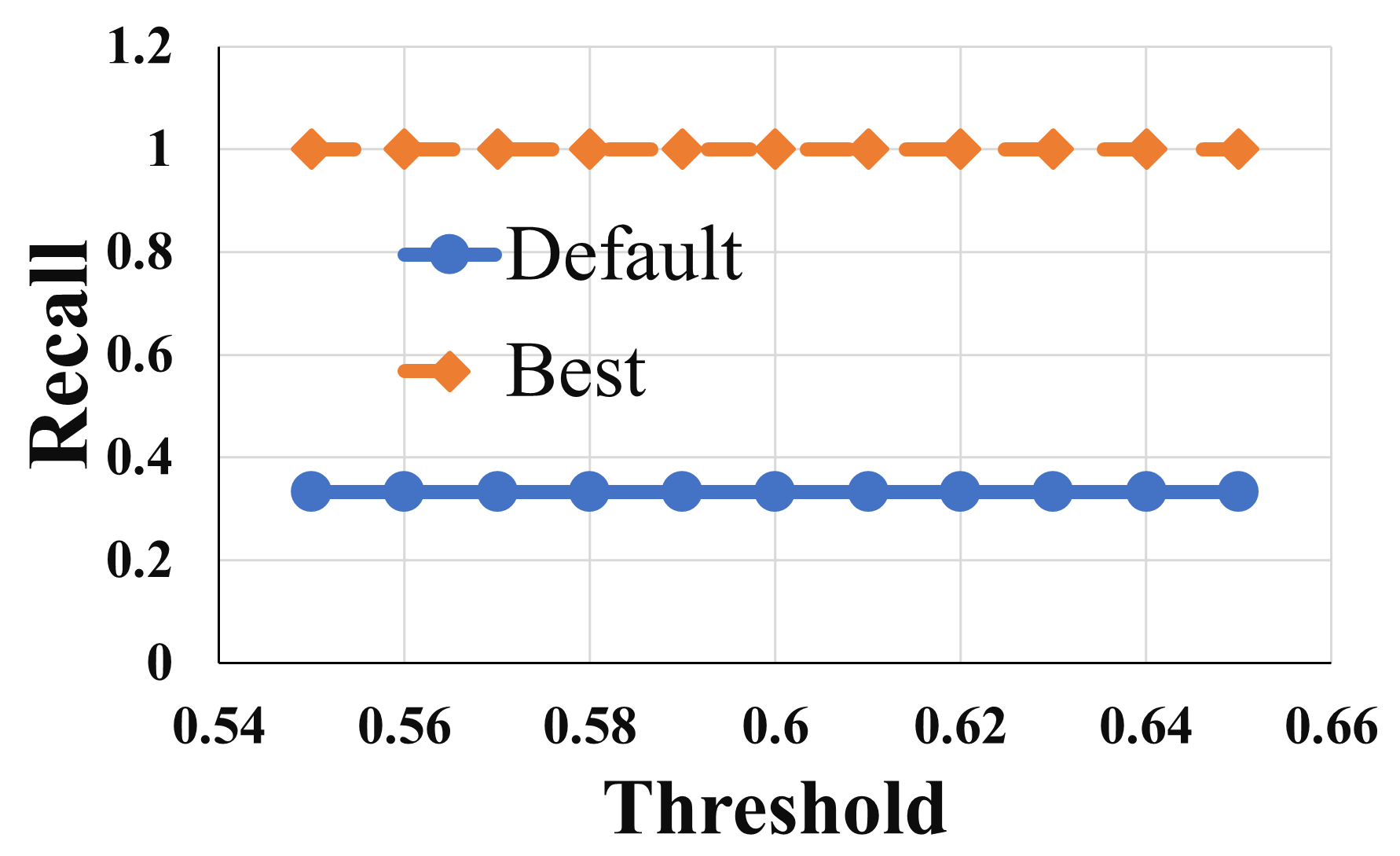}
    \caption{NIGHT}
    \label{fig:recall_night}
\end{subfigure}
    \vspace{-0.2in}
 \caption{Parameter tuning impact for Face-recognition.}
  \label{fig:recall-phy}
 \vspace{-0.1in}
\end{figure}
{\bf Face-recognition VAP:} In this experiment, we place face cutouts
of 10 unique individuals in front of the camera as a fixed static
scene and evaluate the performance of the most accurate
face-recognition AU, Neoface-v3~\cite{NIST}\footnote{This
  face-recognition AU is ranked first in the world in the most recent
  face-recognition technology benchmarking by NIST.}, for various
camera settings and for different face matching thresholds.  Since
this face-recognition AU has high precision despite environmental
changes, we focus on measuring Recall, \ie true-positive rate.

We rerun this experiment for two environmental conditions, \ie {DAY}
and {NIGHT} conditions in our lab, simulated using two sources of
light. One of them is always kept \emph{ON}, while the other light
source is manually turned {ON} or {OFF} to simulate DAY and NIGHT
environmental conditions, respectively.

We compare AU results under the manufacturer-provided ``Default"
camera settings and ``Best'' settings for the four camera parameters.
To find the ``Best'' settings, we exhaustively change the four camera parameters to
find the setting that gives the highest Recall value.  Specifically,
we vary each parameter from 0 to 100 in steps of 10 and capture the
frame for each camera setting. This gives us $\approx$14.6K (11$^4$)
frames for each condition. Changing one camera setting through the
VAPIX API takes about 200ms, and in total it took about \emph{7 hours}
to capture and process the frames for each condition.

~\figref{fig:recall_day} shows the Recall for the {DAY} condition
for various thresholds  ~\figref{fig:recall_night} shows the recall
for the night condition for various thresholds.
We see that
(1) under the ``default''
setting, the recall for the {day} condition goes down at higher thresholds,
indicating that some faces were not recognized, whereas for
the {night} condition, the recall remains constant at a low value for
all thresholds, indicating that some faces were not being recognized
regardless of the face matching thresholds.
(2) in contrast, when we changed the camera parameters for both conditions
to the ``Best'' settings, the AU achieves the highest Recall (100\%),
confirming that all the faces are correctly recognized. 
(3) The ``Best'' settings for {DAY} condition ([80,80,60,40]) is different from that for
{NIGHT} condition ([100,90,30,70]).

}

\vspace{-0.05in}
\input{cam-vs-au}

%% file: cam-vs-au.tex
\subsection{Optimal camera setting is AU-specific}
\label{subsec:cam-vs-au}

\begin{figure}[tb]
\begin{subfigure}[]{0.5\linewidth}
\centering
    \includegraphics[height=0.8 in]{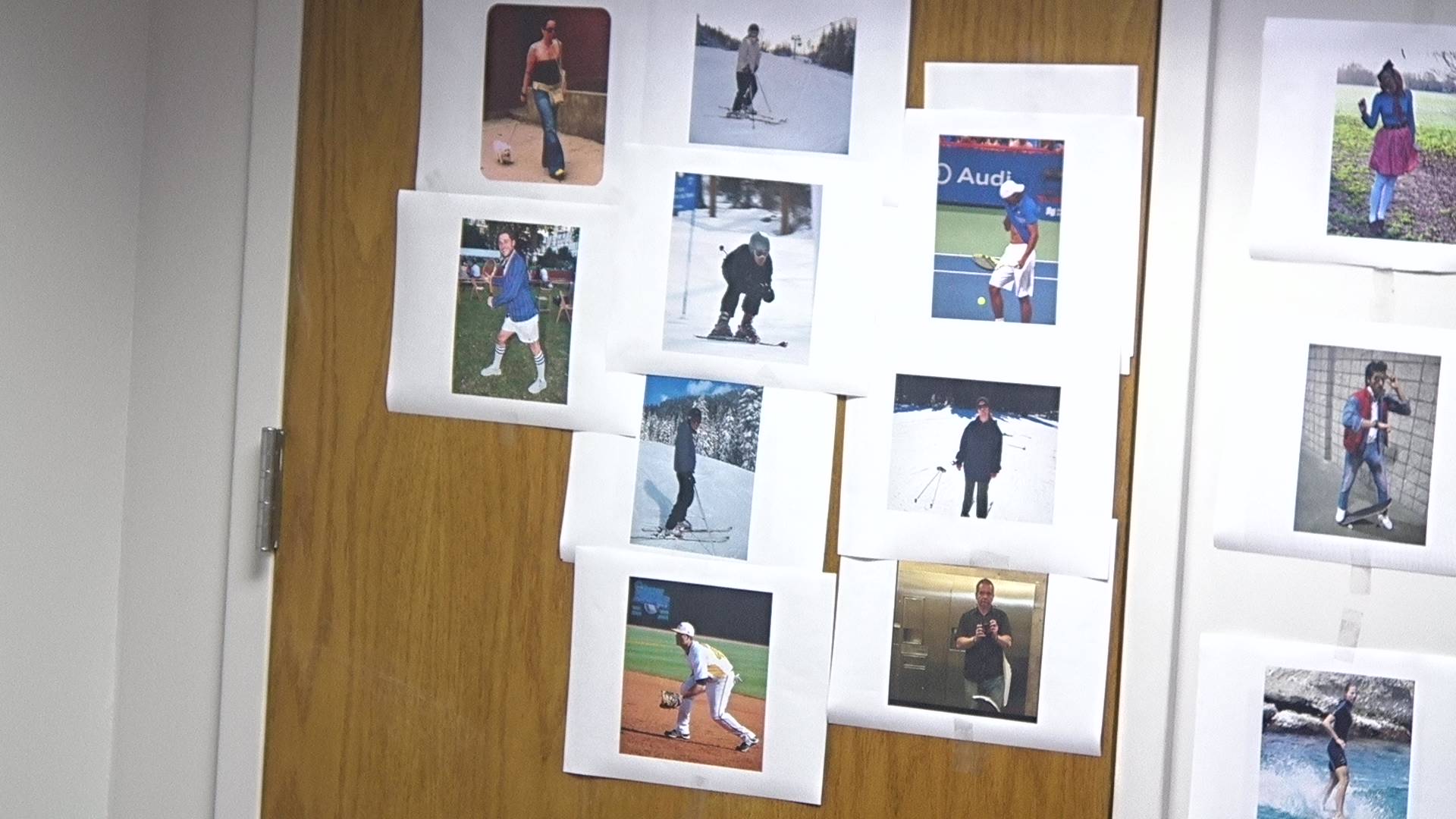}
    \caption{DAY}
    \label{fig:day_capture}
\end{subfigure}%
\begin{subfigure}[]{0.5\linewidth}
\centering
    \includegraphics[height=0.8 in]{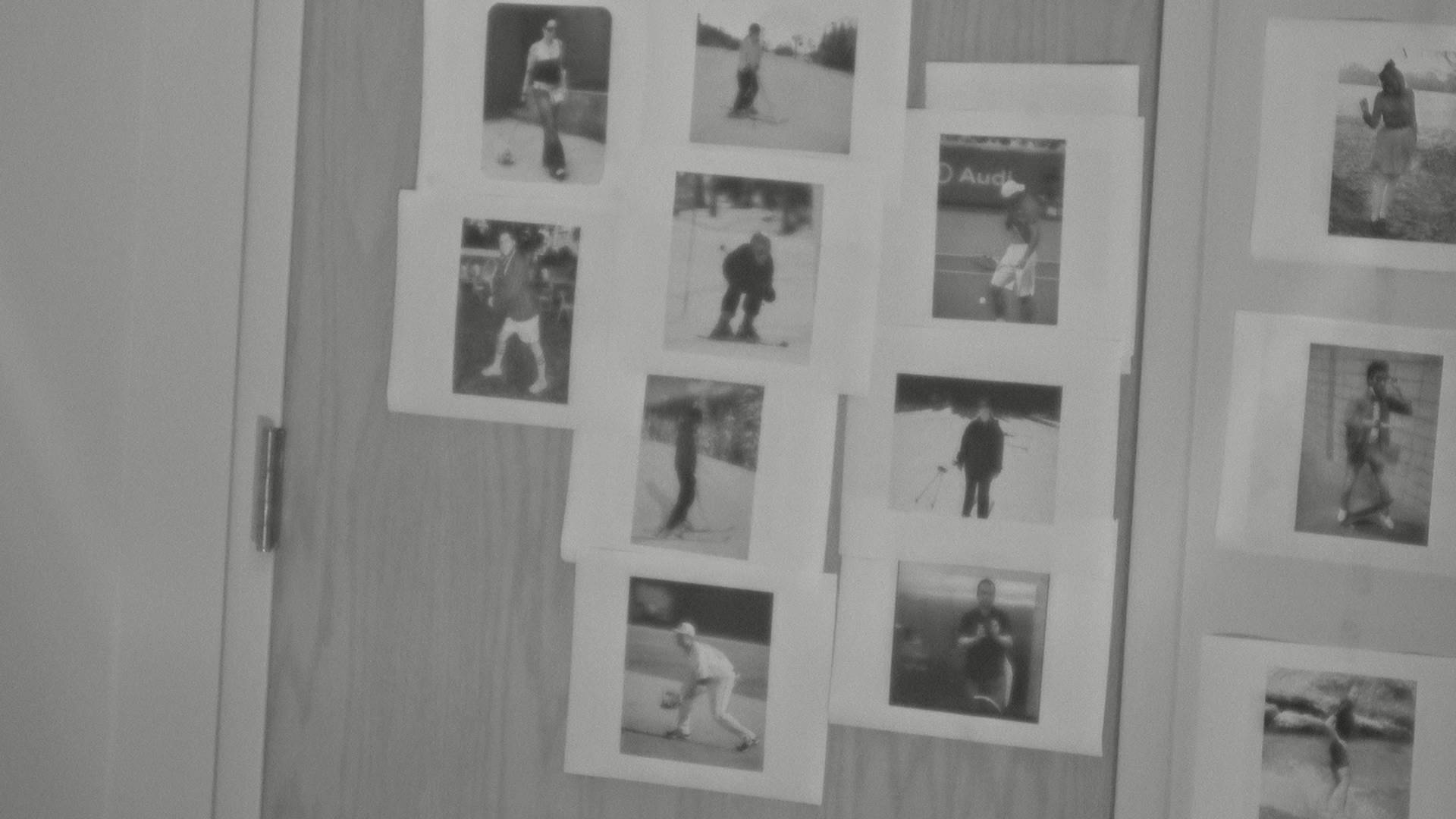}
    \caption{NIGHT}
    \label{fig:night_capture}
\end{subfigure}
 \caption{Camera captures under different environment and visual impact.}
  \label{fig:day-night-capture}
 \vspace{-0.2in}
\end{figure}

Along with the environment, to observe the impact of camera
  parameters on various AUs, we printed 12 different person
  cutouts obtained from COCO dataset~\cite{lin2014microsoft} and
  placed them in front of an Axis network camera. we use
  Efficientdet~\cite{tan2020efficientdet} as person-detection AU and
  RetinaNet~\cite{deng2019retinaface} as face-detection AU and observe
  the impact on each of these AUs individually under \emph{DAY} and
  \emph{NIGHT} condition simulated inside our lab using two light
  sources. One of them is always kept \emph{ON}, while the other light
  source is manually turned {ON} or {OFF} to simulate DAY and NIGHT
  environmental conditions, respectively.
For each of these conditions, we vary the four image appearance camera
parameters, \ie brightness, contrast, sharpness and color-saturation
ranging from 0 to 100 at a step of 10. ~\figref{fig:day-night-capture}
shows the images captured under the default camera setting for \emph{DAY}
and \emph{NIGHT} condition. To find the ``Best'' settings for a
specific AU, we change the four camera parameters to find the setting
that gives the \emph{highest mAP}.  Specifically, we vary each
parameter from 0 to 100 in steps of 10 and capture the frame for each
camera setting. This gives us $\approx$14.6K (11$^4$) frames for each
condition. Changing the camera setting through the VAPIX API takes
about 200ms, and in total it took about \emph{7 hours} to capture and
process the frames for each condition.

Table~\ref{tab:day-night-best-cam} shows that the best I-A camera parameter setting for different AUs are unique. Furthermore, these \emph{Best} camera settings not only vary across different AUs but change due to environmental condition changes (i.e., from DAY to NIGHT), also shown in ~\tabref{tab:day-night-best-cam}. This motivates the need for capturing AU specific perception in tuning the camera parameters.

\begin{table}[h!]
\centering
\caption{Best settings across different AUs for various env.}
\label{tab:day-night-best-cam}
{\small
\begin{tabular}{|c|c|c|}
    \hline
    AU-best &  \multicolumn{2}{|c|}{Best camera setting}\\
    & \multicolumn{2}{|c|}{[brightness, contrast, color, sharpness]}\\
    \cline{2-3}
    & DAY & NIGHT \\
    \hline
    Person Detection-best & [80,90,70,100] & [40,90,60,100] \\
    \hline
    Face Detection-best & [80,90,60,80] & [60,40,90,90] \\
    \hline
\end{tabular}
}
\vspace{-0.1in}
\end{table}

%% file: challenges.tex
\section{Challenges and Approaches}
\label{section:challenge}


Designing \approach to automatically tune camera parameter settings to enhance video analytics accuracy faces several
challenges. In this section, we discuss these challenges and our
approaches to address each one of them.

{\bf Challenge 1: Identifying the best camera setting for a particular
  scene.} Cameras deployed across different locations observe
different scenes. Moreover, the scene observed by a particular camera
at any one location keeps changing based on the environmental
conditions, lighting conditions, movement of objects in the field of
view, etc. In such a dynamic environment, how can we identify the best
camera setting that will give the highest AU accuracy for a
particular scene? The straightforward approach of collecting
data for all possible scenes that can ever be observed by the camera and
training a model that gives the best camera settings for a given
scene is infeasible.
\cut{
 However, it is impossible to know apriori the various scenes
that would be observed by the camera and training such a model is not
practical.
}

\textbf{Approach.} To address this challenge, we propose to
use an online learning method. Particularly, we use Reinforcement
Learning (RL) ~\cite{introduction-to-rl}, in which the agent learns
the best camera settings on the go. 
\cut{
In this way, we do not have to know
a priori the various scenes that the camera would observe. Instead, the
RL agent learns and identifies automatically the best camera
settings that give the highest AU accuracy for any particular
scene.
}
Out of several recent RL algorithms,
we choose 
the SARSA~\cite{q-lambda} RL algorithm for identifying the best
camera settings (more details provided in
\S\ref{camera-parameters-tuning}).

{\bf Challenge 2: No Ground truth in real-time.}
Implementing online RL requires knowing the reward/penalty for every action
taken during exploration and exploitation, \ie what effect 
a particular camera parameter setting will have on the accuracy change of
the AU.  Since no ground truth of the AU task, \eg face detection, is
available during normal operation of the real-time video analytics
system, detecting a change in accuracy of the AU during runtime is
challenging.

\textbf{Approach.} 
We propose to \textit{estimate} the accuracy of the AU. Each AU,
depending on its function has a preferred method of measuring
accuracy, \eg for face detection AU, a combination of mAP and
true-positive IoU is used, whereas for face recognition AU,
the true-positive match score is used. Accordingly, we propose to have
a separate {estimator} for each AU.
\cut{, that estimates the quality
of analytics for that specific AU, depending on its function.
}
We design such \textit{AU-specific analytics quality estimators} to be 
light-weight so that they can be used by the RL agent in real-time
(more details provided in \S\ref{sec:AU-specific}).





{\bf Challenge 3: Extremely slow initial RL training.} 
%
Online learning at each camera deployment setup requires initial RL
training, which can potentially take a very long time for two key
reasons: (1) Capturing the environmental condition changes such as the
time-of-the-day effect requires waiting for the Sun's movement through
the entire day until night, and capturing weather changes requires
waiting for weather changes to actually happen. (2) Taking an action
on the real camera, \ie changing the camera parameter setting, incurs a
significant delay of about 200 ms. This delay fundamentally limits the
speed of state transition and hence the learning speed of RL to only 5
actions per second.


\textbf{Approach.}  In order to speed up the initial RL training, we
propose a novel concept called {\em Virtual Camera (VC)}. A VC mimics
the effect of environmental conditions and camera setting changes on
the frame capture of a real camera. This has two immediate
benefits. {First}, it can effectively complete an action of
``camera setting change'' almost instantaneously.
{Second}, it can
augment a single frame captured by the real camera with many new
transformed frames as if they were captured by the real camera under
different conditions. Together, these two benefits allow the RL system
to explore an order of magnitude more states and actions per unit time
(more details provided in \S\ref{subsec:VC}).

%% file: design.tex
\section{$\approach$ Design}
\label{section:design}
\begin{figure}[tp]
   \centering \includegraphics[width=0.95\linewidth]{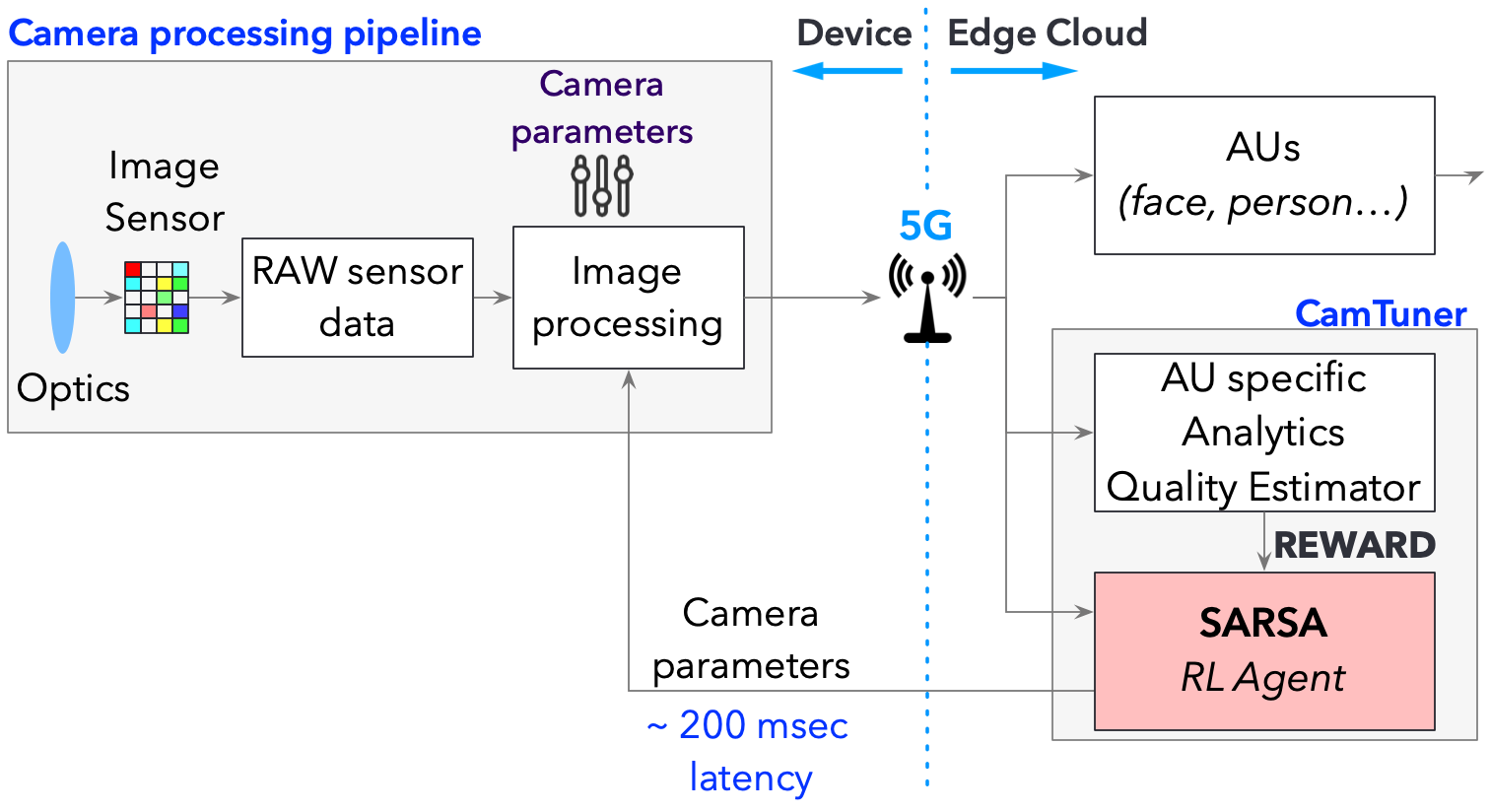} 
   \caption{$\approach$ system design.} 
   \vspace{-0.2in}
\label{fig:system_design}
\end{figure}

\figref{fig:system_design} shows the system-level architecture for
\approach, which automatically and dynamically tunes the camera
parameters to enhance the accuracy of AUs in the VAP.
\approach\ augments a standard VAP shown in Figure~\ref{fig:pipeline}
with two key components: a Reinforcement Learning (RL) engine, and
an AU-specific analytics quality estimator.  In addition, it
employs a third component, a Virtual Camera (VC), for fast initial RL
training.

\cut{Next, we describe these three components in detail.
}

\vspace{-0.05in}
\input{RL}

\vspace{-0.1in}
\input{AU-specific}

\vspace{-0.05in}
\input{VC}

%% file: RL.tex
\subsection{Reinforcement Learning (RL) Engine}
\label{camera-parameters-tuning}

The RL engine is the heart of \approach\ system, as it is the one that automatically chooses the best camera settings for a particular
scene. 
Q-learning~\cite{q-learning} and SARSA~\cite{q-lambda} are two popular RL algorithms that are quite effective in learning the best action to take in order to maximize the reward. We compared these two algorithms and found that training with SARSA achieves slightly faster 
convergence and also slightly better accuracy than with Q-learning. Therefore, we use SARSA RL algorithm in \approach.


SARSA is similar to other RL algorithms. An
agent interacts with the environment (\textit{state}) it is in,
by taking different \textit{actions}. As the agent takes actions, it
moves into a new state or environment. For each action, there is an
associated \textit{reward} or penalty, depending on whether the new
state is desirable or not. Over a period of time, as the agent
continues taking actions and receiving rewards and penalties, it
learns to maximize the rewards by taking the right actions,
which ultimately lead the agent towards desirable states.



SARSA does not require any labeled data or pre-trained model, but it does require a clear definition of the \textit{state}, \textit{action} and \textit{reward} for the RL agent. This combination of \textit{state}, \textit{action} and \textit{reward} is unique for each application and needs to be carefully chosen, so that the agent learns exactly what is desired. In our setup, we define them as follows:


\textit{\underline{State}}: A state is a tuple of two vectors, $s_t=<P_{t}, M_{t}>$,
where $P_t$ consists of the
current brightness, contrast, sharpness, and color-saturation parameter values on
the camera,
and $M_{t}$ consists of the measured values of 
brightness,
contrast, 
color-saturation, and
sharpness
of the captured frame at time $t$,
measured as in~\cite{bezryadin2007brightness,peli1990contrast,hasler2003colormeasuring, de2013sharpness}.

\textit{\underline{Action}}: The set of actions that the agent can
take are (a) increase or decrease one of the brightness, contrast,
sharpness or color-saturation parameter value, or (b) not change any parameter values.

\textit{\underline{Reward}}: We use an AU-specific analytics quality
estimator as the immediate reward function (r) for the SARSA
algorithm. 
Along with considering immediate reward, the agent
  also factors in future reward that may accrue as a result of the
  current actions. Based on this, a value, termed as Q-value (also
  denoted as $Q(s_t,a_t)$) is calculated for taking an action $a_t$ when in
  state $s_t$ using Equation~\ref{eq:sarsa}.
\begin{equation}
\vspace{-0.05in}
\label{eq:sarsa}
Q(s_t,a_t) \leftarrow Q(s_t,a_t) +
\alpha \left[ r + \gamma \cdot Q(s_{t+1}, a_{t+1}) - Q(s_t,a_t) \right]
\end{equation}
Here, $\alpha$ is learning rate (a constant between 0 and 1) used to control how
much importance is to be given to new information obtained by the
agent. A value of 1 will give high importance to the new information
while a value of 0 will stop the learning phase for the agent.

Similar to $\alpha$, $\gamma$ (also known as
the discount factor) is another constant used to control the importance given by the agent
to any long term rewards. A value of 1 will give very high importance
to long term rewards while a value of 0 will make the agent ignore any
long term rewards and focus only on the immediate rewards. If the environmental conditions change very frequently, a lower value, \eg 0.1, can be
assigned to $\gamma$ to prioritize immediate rewards, while if
the conditions do not change frequently, a higher value, \eg 0.9,
can be assigned to prioritize long term rewards.

{\bf Exploration vs. Exploitation.}
We define a constant called $\epsilon$ (between 0 and
1) to control the balance between exploration vs. exploitation in taking actions.
At each step, the agent generates a
random number between 0 and 1; if the random number is greater than
the set value of $\epsilon$, then a random action (exploration) is chosen.

\cut{
  If the quality estimate improves, it is considered as a reward,
  whereas if the estimate decreases, it is considered as a penalty.  }

%% file: AU-specific.tex
\subsection{AU-specific Analytics Quality Estimator}
\label{sec:AU-specific}


In online operations,
the RL engine 
needs to know whether its actions are changing the AU accuracy in the
positive or negative direction. In the absence of ground truth, the
{\em analytics quality estimator} acts as a guide and generates the
reward/penalty for the RL agent. 

{\bf Challenges.} There 
are three key challenges in designing an online analytics quality {estimator}.
\textbf{(1)} During runtime, AU quality
estimation has to be done quickly, which implies a model that is small in
size. \textbf{(2)} A small model size implies using a shallow neural
network. For such a network, what representative features should
the {estimator} extract that will have the most impact on the
accuracy of AU output?
\textbf{(3)} Since different types of AUs (\eg
face detector, person detector) perceive the same representative
features differently, the {estimator} needs to be AU-specific.


{\bf Insights.} 
We make the following observations about estimating the quality of
AUs.
$\textbf{(1)}$ Though estimating the precise accuracy of AU on a frame
requires a deep neural network,
estimating the coarse-grained accuracy, \eg in increments of 1\%,
may only require a shallow neural network.
\textbf{(2)} Most of the ``off-the-shelf'' AUs use convolution and
pooling layers to extract representative local features~\cite{low-level-features}. In
particular, the first few layers in the AUs extract  low-level
features such as edges, shapes, or stretched patterns that affect the
accuracy of the AU results. We can reuse the first few layers
of these AUs in our estimator to capture the low-level
features. 
\textbf{(3)} To capture different AU perceptions from the same
representative features extracted in the early layers, 
we need to design and train the last few layers of each
quality estimator to be AU-specific. During training, we
need to use AU-specific quality labels.

\begin{figure}[tp]
   \centering \includegraphics[width=0.7\linewidth]{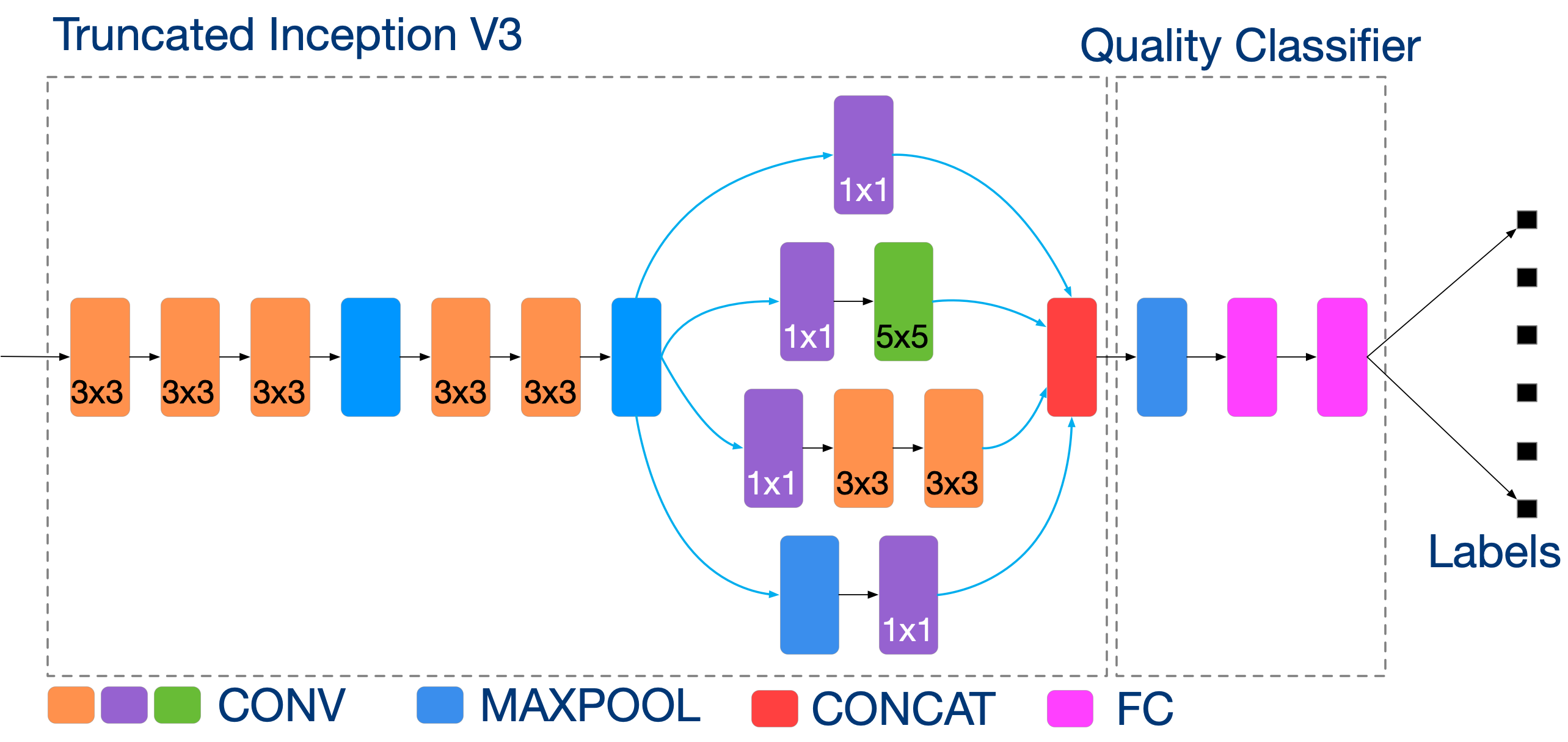} \caption{AU-specific analytics quality estimator design.} \label{fig:quality_estimator_design}
   \vspace{-0.2in}
\end{figure}

{\bf Design.} 
Motivated by the above insights,
we design our light-weight 
AU-specific analytical quality estimator
to consist of two components: (1) feature
extractor and (2) quality classifier, as shown in ~\figref{fig:quality_estimator_design}.
We use supervised learning to train the AU-specific quality
estimator.

{\em Feature Extractor. } Different AUs and environmental
conditions can manipulate local features of an input frame at
different granularities~\cite{GUPTA2021100057}. For example, blur (\ie\ motion
or defocus blur) affects fine textures while light exposure 
affects coarse textures. While face detector and face recognition AUs
focus on finer face details, person detector is coarse-grained and
it only detects the bounding box of a person. Similarly, in convolution
layers, larger filter sizes focus on global features while stacked
convolution layers extract fine-grained features. To accommodate such
diverse notions of granularities, we use the Inception module from the
Inception-v3 network~\cite{inception-v3}, which has convolution layers with
diverse filter sizes. 
\cut{
In particular, the feature extractor of our
quality estimator is built using all the layers in Inception-v3 until
the first inception module.
}

{\em Quality classifier.} The goal of the quality classifier is to
take the features extracted by the feature extractor and estimate the
coarse-grained accuracy of the AU on an input frame, \eg in increments of
1\%. As such, we divide the AU-specific accuracy measure into multiple
coarse-grained labels, \eg from 0\% to 99\%, and use
fully-connected layers whose output nodes generate AU-specific
classification labels. 


Detailed design and training of two concrete AU-specific analytics quality
estimators are described as follows.

{\em (1) Face recognition AU:} 
{The quality classifier of face recognition
consists of 2 fully-connected layer} and has
101 output classes. One of the classes signifies no
match, while the remaining 100 classes correspond to match scores
between 0 to 100\% in units of 1\%.

To generate the labeled data, we used 300 randomly-sampled
celebrities from the celebA dataset~\cite{liu2015faceattributes}.
We choose two images per person. We use one of them as a reference image
and add it to the gallery.
We use the other image to generate multiple
variants by 
applying digital transformations on the image.
These variants ($\sim$4 million) form the query images.
%
For each query image, we obtain the match score (a value
between 0 and 100\%) using the Face recognition AU,
\emph{Neoface-v3}. The query images along with their match score form
the labeled samples, which are used to train the quality estimator.

{\em (2) Face and object detection AU. }  {The quality
  classifier of face  and object (\ie car and person) detection AU consists of 2 fully-connected
  layers}, and has 201 output classes to predict the quality estimate of
  the face and object detection AU for a given frame. One of the classes
  signifies AU cannot detect anything accurately, and the remaining
  200 classes correspond to the cumulative mAP score between 0 to 100
  and IoU score between 0 to 1, \ie $mAP + IOU_{True-Positive}*100$.
To generate the labeled data to train face-detection AU specific quality estimator,
we used the Olympics~\cite{niebles2010modeling} and HMDB~\cite{Kuehne11} datasets,
and created $\sim$7.5 million variants of the video frames by 
applying digital transformations.
Then, for each frame, we use the face detection AU (\ie RetinaNet~\cite{deng2019retinaface}) to determine the analytical quality estimate. Similarly, we use the object detection AU (\ie EfficientDet~\cite{tan2020efficientdet}) on labeled images from COCO dataset~\cite{lin2014microsoft} that contain car and person object classes and their augmented variants. 
%
The video frames/images and their quality estimates form the
labeled samples, which are used to train the estimator model.

For both the classifier training, we use a cross-entropy loss function
to train AU-specific analytics quality estimators, initial learning
rate is $10^{-5}$, and we use Adam Optimizer~\cite{kingma2014adam}

%% file: VC.tex
\subsection{Virtual Camera}
\label{subsec:VC}

\cut{
We design a virtual camera (VC) to accelerate initial RL training by
mimicking the effect of environmental conditions and camera setting changes
on the frame capture of a real camera.
}

\begin{figure}[tb]
\centering
    \includegraphics[width=3.3 in]{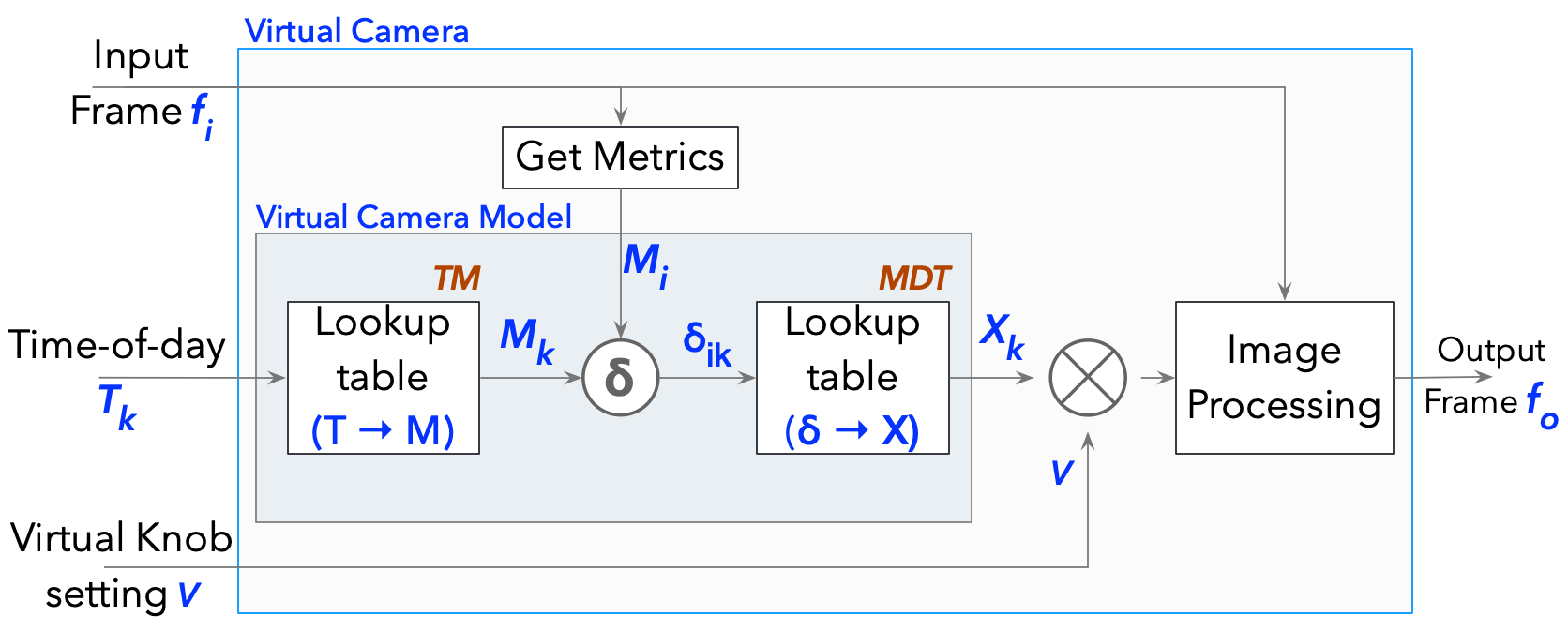}
    \caption{VC block diagram.}
    \label{fig:vc-block-diagram}
\vspace{-0.1in}
\end{figure}
 
{\bf Definition.}  A VC (shown in~\figref{fig:vc-block-diagram}) takes
an input frame $f_i$, captured by a real camera, the target time-of-the-day $T_k$, and VC parameter
settings $V$, as input, and
outputs a frame $f_o$ as if it was captured by the physical camera at
time $T_k$. To generate a frame at time $T_k$, VC uses a composition function $Compose(X_k, V)$, which composes output frame $f_o$ using $X_k$, which is the transformation that augments
the environmental effects corresponding to the target time $T_k$ on input frame $f_i$, and $V$, which is the VC parameter settings. The composition function is defined as $X_k * 10^{V - 0.5}$, which considers $X_k$ and $V$ simultaneously, similar to a real camera. Using this composition function, $X_k$ is scaled up if the value of $V$ is greater than 0.5 and scaled down if the value is less than 0.5; no scaling of $X_k$ happens for $V$ equal to 0.5.

%


To understand how VC works, we first introduce an important 
definition. Each frame $f_i$, from a real physical camera, possesses distinct values of brightness, contrast, colorfulness and sharpness metrics, denoted as a {\em metric (or feature) tuple}
$M_i = <\alpha_M$, $\beta_M$, $\gamma_M$, $\zeta_M>$.
The unique metric tuple encapsulates
the environmental conditions and the default physical camera settings
when the frame was captured.
\cut{
For example, at two different times of the day
and hence under two environmental conditions,
a surveillance camera using the default settings
would output two frames with different metric tuples.
}





\begin{figure}[tp]
\begin{subfigure}[]{0.48\linewidth}
\centering
    \includegraphics[height=0.9 in]{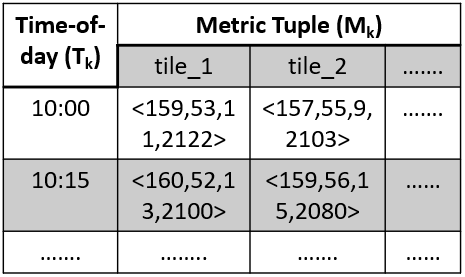}
    \caption{TM table}
    \label{fig:vc-table}
\end{subfigure}%
\begin{subfigure}[]{0.48\linewidth}
\centering
    \includegraphics[height=0.9 in]{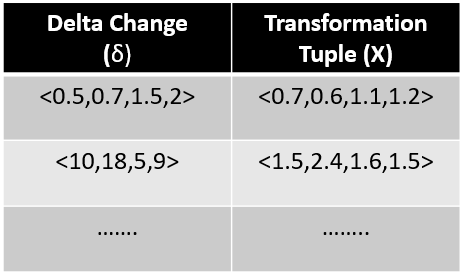}
    \caption{MDT Table}
    \label{fig:delta-to-config}
\end{subfigure}
\vspace{-0.1in}
 \caption{Offline generated tables for VC.}
  \label{fig:vc-tables}
 \vspace{-0.2in}
\end{figure}

\textbf{Offline profiling phase}: 
VC derives two tables for a given physical camera deployment during an offline profiling
phase and then uses the two tables 
during online operation to generate the output frame $f_o$. 

The first table (TM) maps a given time-of-the-day $T_k$ to the metric tuple $M_k$
which captures the distinct values of brightness, contrast,
colorfulness and sharpness metrics of frames taken
by the physical camera with the default settings at time $T_k$.
We generate the table to cover the full 24-hour period
with a granularity of 15 minutes, \ie the table has one
mapping for every 15 minutes, for a total of 96 mappings.
To construct the table, we use a full 24-hour long video and break it
into 15-minute video snippets.
%
We extract all the frames from the video snippet for each 15-minute
interval $T_k$. We divide each frame into 12 tiles, obtain the
corresponding metric tuple for each tile, and compute the mean metric
tuple for the corresponding tiles in all frames in the 15-minute
interval as the metric tuple for that tile,
and the list of tuples for all 12 tiles form the entry 
for time $T_k$ in the table, as shown in~\figref{fig:vc-table}.


The second table (MDT) maps the difference between two metric tuples $M_i$
and $M_k$, $\delta(M_i, M_k)$,
to the corresponding transformation tuple $X_k$ that would effectively transform a frame captured by the
physical camera with metric tuple $M_i$ to become a frame captured by
the physical camera for the same scene with metric tuple $M_k$.
We note since each camera parameter can take 11 values, from 0 to 100 with
increments of 10, the difference between any two metric tuples can possibly be mapped to one of these 14K ($11^4$) settings.
We construct the entries for the table backward as follows.
(1) We select a random frame from each 15-minute interval
to form a collection of 96 frames with varying environmental conditions,
\ie corresponding to different time-of-the-day.
(2) For each possible transformation $X_k$, 
we transform the 96 frames into 96 virtual frames. We then obtain the delta
metric tuples between each pair of original and 
transformed frames, calculate the median
of the 96 delta metric tuples, $\delta_k$, and store the pair of 
$<\delta_k, X_k>$ in the table.
(3) We repeat the above process for all possible transformation settings
(14K in total) 
to populate the table, as shown in~\figref{fig:delta-to-config}.

Finally, at runtime when the table is used by the VC, if 
the entry for a given delta metric tuple $\delta_i$ is empty,
we return the entry whose delta metric tuple $\delta_k$ is closest
to $\delta_i$ using L1-norm.

\cut{

 \begin{figure}[tp]
\begin{subfigure}[]{0.5\linewidth}
\centering
    \includegraphics[height=1.2 in]{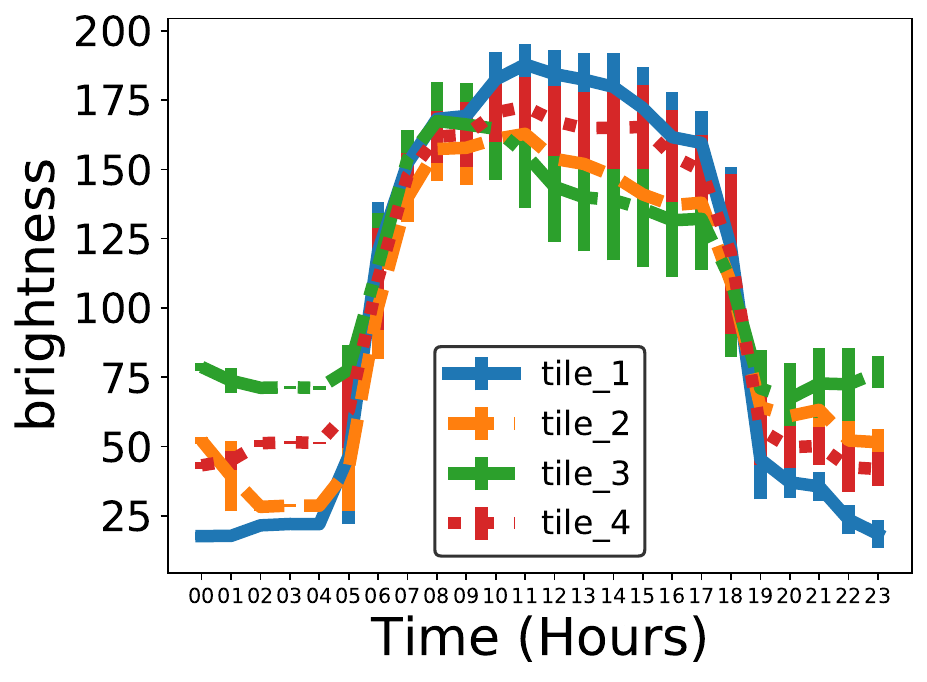}
    \caption{Brightness}
    \label{fig:vc-bright}
\end{subfigure}%
\begin{subfigure}[]{0.5\linewidth}
\centering
    \includegraphics[height=1.2 in]{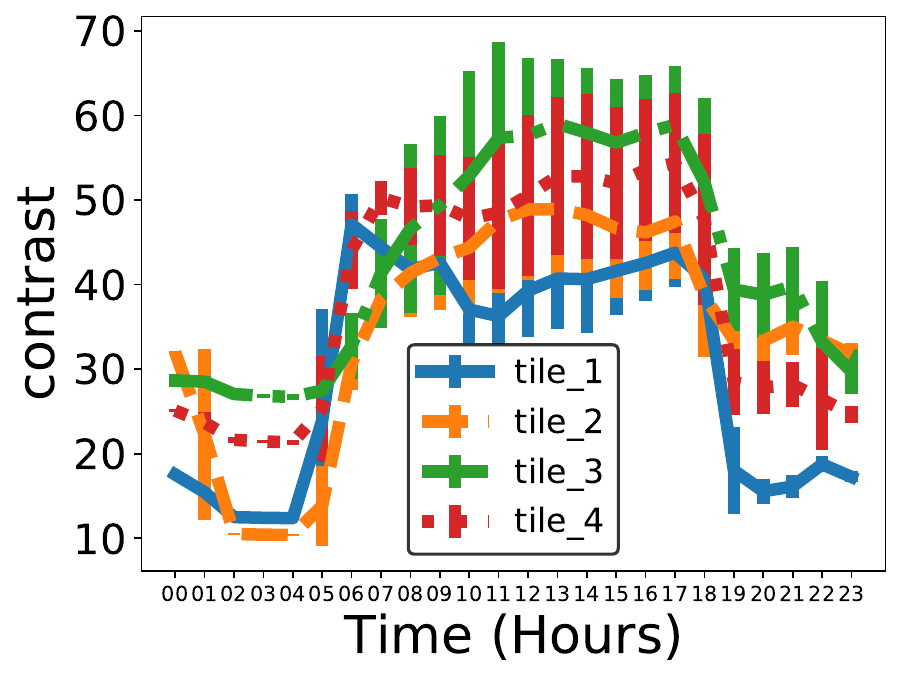}
    \caption{Contrast}
    \label{fig:vc-contrast}
\end{subfigure}
\begin{subfigure}[]{0.5\linewidth}
\centering
    \includegraphics[height=1.2 in]{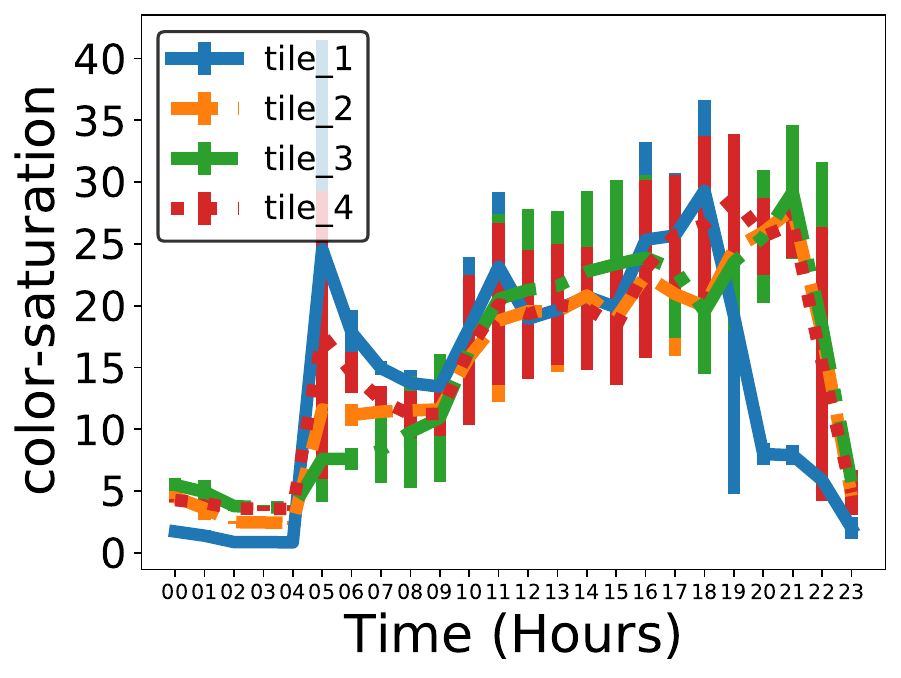}
    \caption{Color-Saturation}
    \label{fig:vc-color}
\end{subfigure}%
\begin{subfigure}[]{0.5\linewidth}
\centering
    \includegraphics[height=1.2 in]{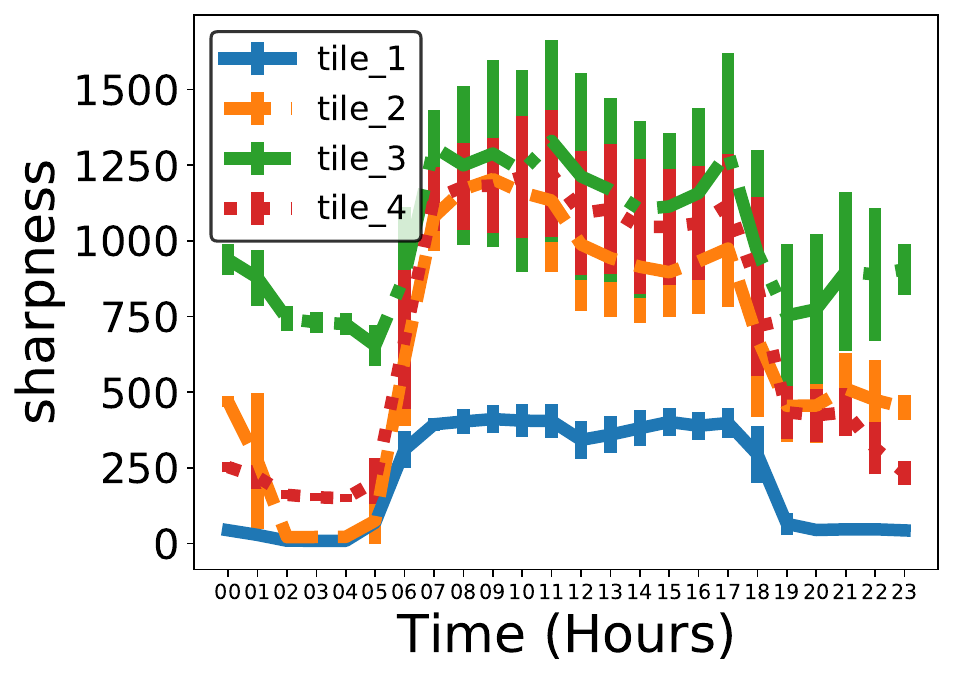}
    \caption{Sharpness}
    \label{fig:vc-sharp}
\end{subfigure}
\vspace{-0.2in}
 \caption{Day-long feature profiles for different tiles \comment{can be removed}}
  \label{fig:vc-profile}
\end{figure}
}

\textbf{Online phase.} VC transforms the input frame $f_i$ to output frame $f_o$ in five steps.
(1) It measures the current metric  tuple
$M_i = <\alpha_M$, $\beta_M$, $\gamma_M$, $\zeta_M>_{curr}$ from input frame $f_i$;
(2) It looks up 
\emph{Time-to-Metric (TM)} table 
for the metric tuple $M_{k} = <\alpha_M$, $\beta_M$, $\gamma_M$, $\zeta_M>_{desired}$ 
that corresponds to the target time of the day ($T_k$); 
(3) It calculates the difference between $M_i$ and $M_k$, $\delta(M_i, M_k)$ or $\delta_{ik}$;
(4) It looks up \emph{Metric-difference-to-Transformation (MDT)} table to find the transformation
$X_k = <\alpha_{X}$, $\beta_{X}$, $\gamma_{X}$, $\zeta_{X}>_{applied}$ 
that corresponds to $\delta_{ik}$; 
(5) It applies $X_k$ along with $V$ using the composition function $Compose(X_k, V)$ 
to input frame $f_i$ and generates output frame $f_o$.

Since different parts of an input frame may exhibit 
varying local feature or metric values,
to improve the effectiveness of virtual knob transformation,
instead of applying the above steps directly to input frame $f_i$, 
we split it into 12 (3 $X$ 4) equal-sized tiles,
apply Steps 1-3 to 
each of the 12 tiles,
\ie each of $M_i$, $M_k$, and $\delta_{ik}$ consists of 
12 sub-tuples corresponding to the 12 tiles, respectively.
The 12 sub-tuples in $\delta_{ik}$
are looked up in the MDT table to find 12 transformation tuples.
Finally, to ensure smoothness,
we calculate the mean of these 12 sub-tuples $X_k$,
which is then applied to 
input frame $f_i$.

\cut{
Due to page limit, 
construction of the two mapping tables
of VC 
are omitted here and can be found in
Appendix A3.
}

\vspace{-0.1 in}
\subsection{Integrating VC with the RL engine}
\label{subsec:vc-sim}

\cut{
A VC can be used to speed up initial RL training 
upon camera deployment at a new surveillance site,
which enhances the overall SARSA training convergence,
for two reaons:
(1) it can augment each frame in a training video, \eg recorded in an hour
in the morning, with many different frames corresponding
to different environmental conditions, \eg throughout a day, and thus
omit the need to wait for video capture over a long time;
(2) the RL agent can take actions, \ie resetting virtual knob values in the VC,
much faster than driving a physical camera and hence performs 
RL exploration much faster.
}

\cut{
During initial RL training, the RL agent performs
fast exploration by driving the VC as follows.
It reads each frame $f_i$ from the input training video,
and repeats the following exploration steps for 
all time-of-the-day values $T_j$.
At each exploration step $j$,
the agent which is at state $s=<P_{j}, V_j>$ performs tasks:
(1) based on current state (s), 
it takes a random action $a$ and apply that on $V_j$ to get a new virtual knob setting,
$V_{s_{j+1}}$;
(2) it invokes the VC with frame $f_i$, the next time-of-the-day $T_j$, and current VC parameters $V_{s_{j+1}}$ as input,
and the VC outputs frame $f_o$.
The measured tuple $P_{j+1}$ of  brightness, contrast, colorfulness and sharpness metric
values
of output frame $f_o$ 
along with the virtual knob setting $V_{s_{j+1}}$, 
form the new state of the RL agent, $s_{new} = <P_{j+1}, V_{s_{j+1}}>$;
(3) it calculates the reward/penalty by feeding $f_o$ into
the AU-specific quality estimator; and
(4) it updates the Q-table entry $Q(s, a)$.
}

During initial RL training, the RL agent performs
{\em fast exploration} by leveraging VC as follows.
It reads each frame $f_i$ from the input training video,
and repeats the following exploration steps for 
all time-of-the-day values $T_k$.
At each exploration step $j$,
the agent which is at state $s=<P_{j}, M_{j}>$ performs tasks:
(1) based on current state (s), 
it takes a random action $a$ and apply that on $V_j$, which is VC equivalent of $P_{j}$ for a real camera, to get a new virtual knob setting for next exploration step ($j+1$),
$V_{j+1}$;
(2) it invokes the VC with frame $f_i$ for the target time-of-the-day $T_k$, and current VC parameters $V_{j+1}$ as input,
and the VC outputs frame $f_o$.
The measured tuple $M_{j+1}$ of  brightness, contrast, colorfulness and sharpness metric
values
of output frame $f_o$ 
along with the virtual knob setting $V_{j+1}$, 
form the new state of the RL agent, $s_{new} = <V_{j+1}, M_{j+1}>$;
(3) it calculates the reward/penalty by feeding $f_o$ into
the AU-specific quality estimator; and
(4) it updates the Q-table entry $Q(s, a)$.

The above initially trained SARSA model with the VC is then deployed
in the real camera for the normal operations of \approach.  First, the
$\epsilon$ value is set to low (0.1) and $\alpha$ is set to
  high (0.85) so that the SARSA RL agent will go through a short
\emph{adaptation} phase, \eg for an hour, by performing primarily
exploration.  Afterward, the $\epsilon$ and $\alpha$ values
  are set to high (0.9) and low (0.15), respectively, so that SARSA
  performs primarily exploitation using the trained model.

%% file: impl.tex
\section{Implementation}
\label{section:impl}

\subsection{Hardware Setup}
\label{subsec:hwsetup}

For the evaluation, we implemented a VAP using an {Axis Q3505 MK II}
network surveillance camera. We run \approach on a low-end {Intel
  NUC box} while face detection and object detection AUs and
initial pre-training with VC run on a high-end edge-server equipped
with Xeon(R) W-2145 CPU and GeForce RTX 2080 GPU. The captured frames
are sent for AU processing on the edge-server over a 5G network with
an average frame uploading latency of 39.7 ms.
\vspace{-0.15in}
\subsection{Software Implementation}
\label{subsec:sw}

We implemented the SARSA RL agent in Python, the light-weight
AU-specific analytics quality estimators 
in pytorch framework which runs as a service using the
ZeroMQ~\cite{zeromq} networking library, and the Virtual Camera in Python
which is trained on the GPU edge server. 
We use PIL~\cite{pil} and
OpenCV~\cite{cv2} for image processing during the offline profiling phase in
VC design and also during offline training of the SARSA RL agent.
We use Axis' \emph{VAPIX} API to change the
camera parameters decided by the SARSA-RL agent as well as to capture 
input frames.

Similar to a real camera, our VC runs continuously during offline SARSA RL training
and streams the output frames on a NATS~\cite{nats} queue at the same
frames-per-second (FPS) with which the video was captured. Each frame
is sent in BSON format which includes the frame number, frame data
(\ie array of bytes), and timestamp. Like a real camera, VC exposes REST APIs that are used to query and change its settings to allow augmenting various environmental effects.


%% file: eval.tex
\section{Evaluation}
\label{section:eval}

We extensively evaluate the effectiveness of \approach by measuring
its impact on AU accuracy improvement in a VAP via controlled
experimental emulation and in a real deployment
(\S\ref{subsec:mainresult} -- \S\ref{subsec:accipreven}). We also
evaluate its system performance (\S\ref{subsec:overhead}) and the efficacy of its two key
components, AU-specific analytics quality estimator and VC (\S\ref{subsec:components}).

\subsection{End-to-end VAP Performance}
\label{subsec:mainresult}

We first evaluate the effectiveness of \approach by 
comparing AU accuracy of five different VAPs.

\subsubsection{Experimental Setup}
\label{subsubsec:expsetup}

We compare three \approach\ variants against two baseline VAPs.
All system variants, including \approach,
only differ in how the four I-A camera parameters are tuned, while keeping
all automatic parameter setting features
turned on and the rest NAUTO parameters at the default values.
(1) \emph{Baseline}: In the Baseline VAP, the I-A camera parameters are not
adapted to any environmental changes.
(2) \emph{Strawman}: The Strawman approach applies a time-of-the-day
heuristic that tunes the four I-A camera parameters based on a human
perception metric. In particular, we use the BRISQUE quality
metric~\cite{brisque_mittal2012no} and exhaustively search for the
best camera parameters for the first few frames in each hour and then
apply the best camera setting found for the remaining frames in that
hour. This exhaustive search of camera settings using initial frames
takes a few minutes and our results show that performing this
adaptation more often than once per hour does not give significant
improvement. 
(3) \emph{\approach-$\beta$}: This variant of \approach only
  uses a few rounds of online exploration (\ie which takes about 1 hour, same as in online
  exploration performed by \approach), \ie the SARSA RL agent does not
  rely on the VC for initial offline exploration. Instead, at the start
  of online exploration, the \approach-$\beta$ framework is initially
  seeded with an empty Q-table.
(4) \emph{\approach-$\alpha$}: This variant of \approach adjusts
the I-A camera setting dynamically by using only the offline trained
SARSA RL agent, {\ie the agent
does not perform any exploration during online operation.}
(5) \emph{\approach}: The complete \approach framework is seeded with
offline trained SARSA RL agent, and then during online operation, the
agent continues exploration initially and then moves towards
exploitation, as described in \S\ref{subsec:vc-sim}.  For \emph{\approach}-{$\beta$}, 
\emph{\approach}-{$\alpha$} and \emph{\approach}, the RL agent
adaptively adjusts the four I-A camera parameters periodically; the time
interval is configurable and we choose it to be \emph{10s}.

{\bf Experimental methodology.} 
Comparing these 5 VAPs in a real-world deployment is difficult because
(1) even with 5 co-located cameras, it is difficult to see the identical
scene from the same angle;
(2) furthermore, in a real-world deployment,
the captured scenes do not have the ground-truth to measure the AU accuracy.
To overcome the above challenge, we
loop a pre-recorded (original) 5-minute video snippet (a customer
video captured at an airport) labeled with ground-truth through VC -- VC is used here not for RL training but for generating
{\em augmented} input videos that emulate different environmental changes
to be fed into the five VAPs.  In particular, we
gradually change the VC model parameters (\ie digital transformations)
to simulate the changes that happen during the day as the Sun changes
its position and finally sets, and we ensure (through manual inspection)
that same ground-truths are carried over in the VC generated videos
from the original video.
We then project these VC-generated videos on a
monitor screen in front of a real camera,
and run each of the five VAPs in turn. 
{We note that the above controlled experimental setup
is the closest approximation to a real-world deployment.
}

\begin{figure}
    \centering
    \begin{subfigure}[t]{0.495\columnwidth}
        \vskip 0pt
        \centering
        \includegraphics[width=0.99\textwidth]{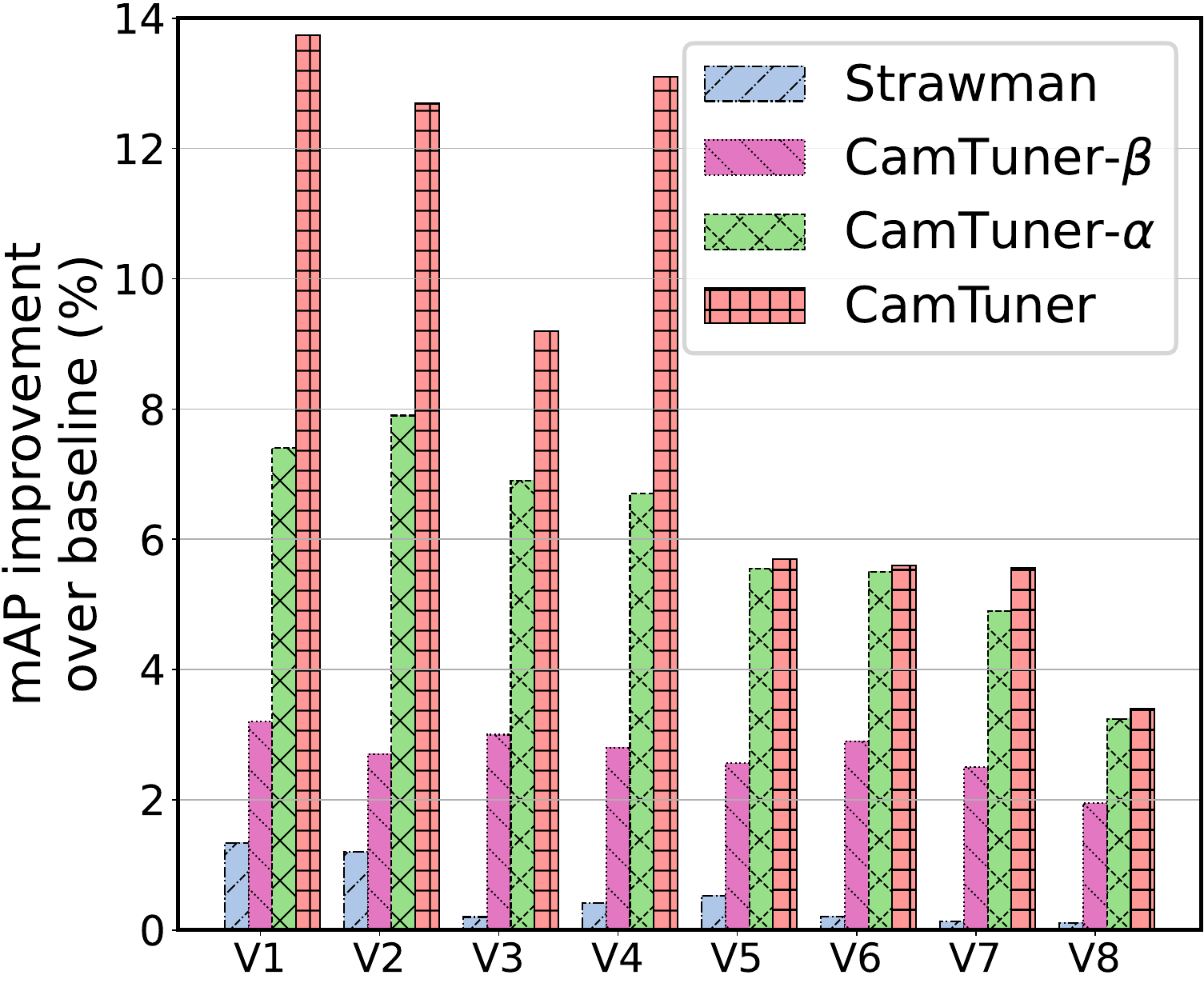}
        \caption{Face detection AU}
        \label{fig:eval_face_det}
    \end{subfigure}
    \hfill
    \begin{subfigure}[t]{0.495\columnwidth}
        \vskip 0pt
        \centering
        \includegraphics[width=0.99\textwidth]{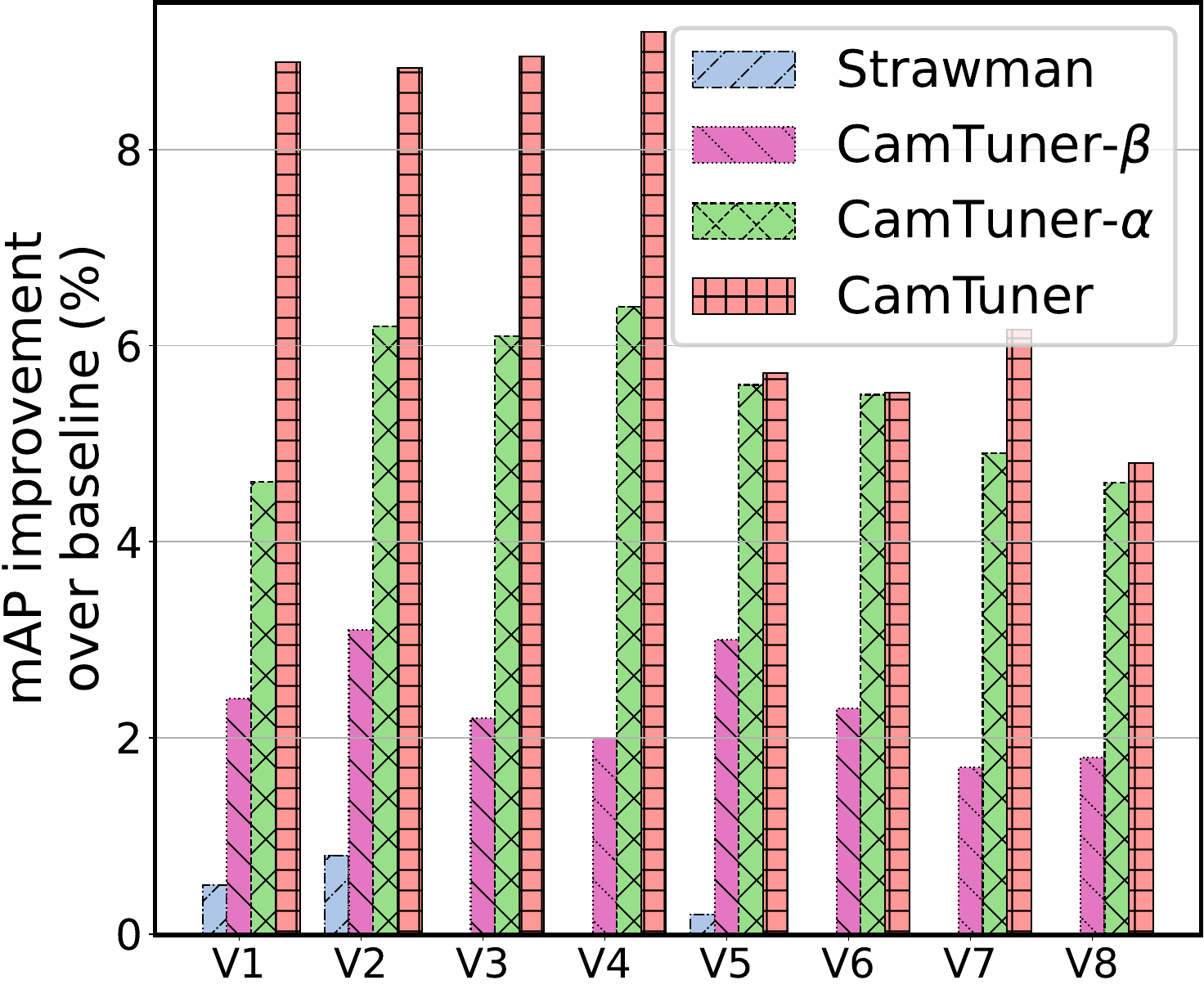}
        \caption{Person detection AU}
        \label{fig:eval_person_det}
    \end{subfigure}
    \vspace{-0.1in}
\caption{mAP improvements for different AUs.}
\label{fig:eval_face_person_au}
\vspace{-0.2in}
\end{figure}


\subsubsection{End-to-end Accuracy}
\label{subsubsec:end2end}
We evaluate the AU accuracy improvement of VAPs 2-5 over VAP 1 for eight
5-minute video segments randomly selected from the VC-generated videos
consisting of 7500 frames each, and the video segments
are separated by 1 hour apart. 
Using the labeled ground-truth,
we evaluate the detection accuracy of the 5 VAPs for
face-detection and person-detection AUs.

~\figref{fig:eval_face_person_au} shows the bar-plot of mAP
improvement of VAPs 2-5 over VAP 1 for the
eight 5-min video segments corresponding to eight different hours of the day.
We make the following observations.
The strawman approach based on the \emph{time-of-the-day} heuristic
can provide only nominal improvement over Baseline, \ie less than
1\% on average across the videos for both face detection and person
detection. Just a few hours of \say{slow} online exploration (\ie with no VC-accelerated offline exploration) enables \emph{\approach}-{$\beta$} to improve face detection accuracy by
       {2.70\%} on average and person detection accuracy by {2.31\%}
       on average over Baseline.
%
In contrast, \emph{fast} offline exploration using virtual camera (with no online exploration) helps
\emph{\approach}-{$\alpha$} to improve face detection accuracy by
{6.01\%} on average and person detection accuracy by {5.49\%} on
average over Baseline.
Finally, dynamically tuning the real camera parameters with online
learning in \approach improves the face detection AU accuracy by up to
13.8\% and person detection AU accuracy by up to 9.2\%, with 
an average improvement of 8.63\% and 8.11\% for face detection AU, and
average improvement of 7.25\% and 7.08\% for person detection AU
compared to Baseline and Strawman, respectively. 

In summary, during offline phase VC helps the SARSA RL agent to quickly train
  through fast and equivalent environmental changes and camera
  parameter changes applied to the input scene. Then
during online operation, a few rounds of exploration helps \approach to achieve better accuracy than directly using the
  initially trained SARSA model with VC (\approach-$\alpha$).

\subsubsection{In-depth Analysis}

\begin{figure}[tb]
\begin{subfigure}[t]{\linewidth}
\centering
    \includegraphics[width=\textwidth]{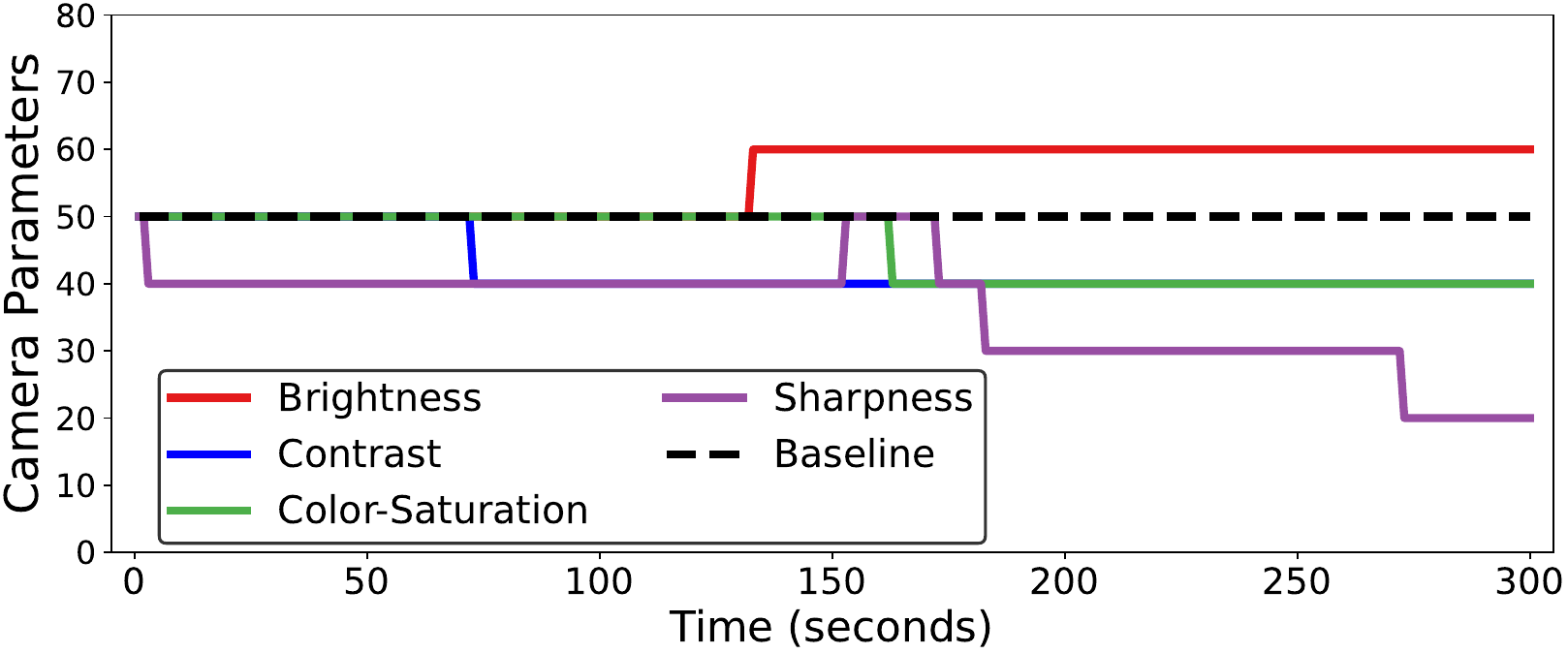}
    \caption{Camera setting adaptation}
    \label{fig:camparam}
\end{subfigure}
\begin{subfigure}[t]{\linewidth}
\centering
    \includegraphics[width=\textwidth]{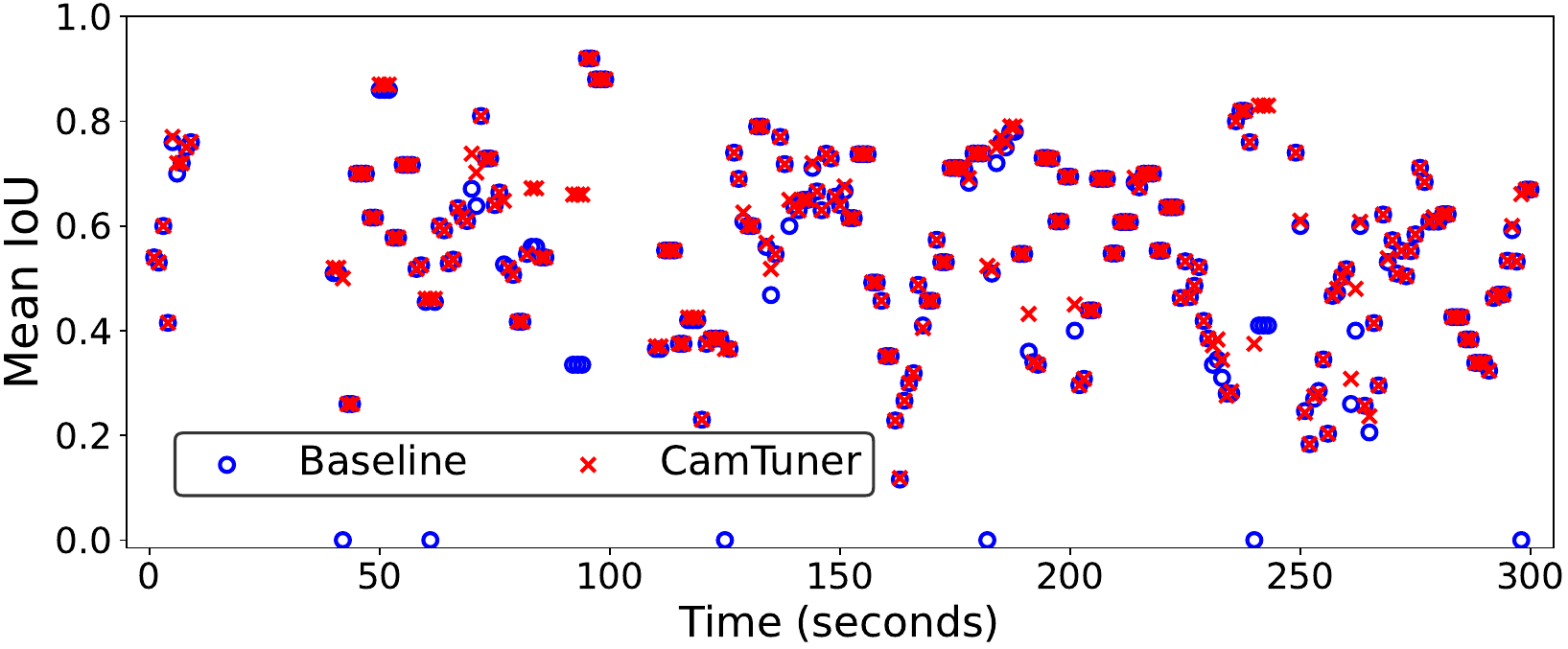}
    \caption{mIoU variation}
    \label{fig:mIoU}
\end{subfigure}%
 \vspace{-0.1in}
 \caption{$\approach$ in operation.}
  \label{fig:camtuner_op}
  \vspace{-0.2in}
\end{figure}

Next, we show how \approach dynamically adjusts the camera parameter setting
for one of the 5-minute video snippet (\ie V3 in~\figref{fig:eval_face_person_au}) used
in \S\ref{subsubsec:end2end} for \textit{face-detection AU}.
Recall at every 10 seconds, 
based on the current environmental condition and
content seen by the camera, \approach
either chooses to ``increment'' or ``decrement'' one of the four parameters,
\ie increase or decrease by 10
within the parameter range of [0, 100],
or keep the previous parameter setting. 
~\figref{fig:camparam} shows how the camera parameters are adapted
throughout the video length during the exploitation phase, and
~\figref{fig:mIoU} shows how the corresponding mean intersection-over-union (mIoU) (\ie
IoU across all ground-truth bounding boxes in each frame) varies
for the \approach-based VAP and the Baseline VAP.

We make the following observations.
(1) Starting with the default camera parameter setting, \ie [50, 50, 50, 50],
\approach decrements the sharpness parameter after looking
into the initial two frames, and then decrements
contrast after 7 tuning intervals (at 70\textsuperscript{th} second).
At the $13^{th}$ tuning interval, it increments a third parameter,
brightness. Then again after two intervals (at 150\textsuperscript{th} second), it increments
the sharpness parameter. In the subsequent interval, \approach decides to
decrement color-saturation after looking into the most recently captured
scene. Finally, \approach further decrements the sharpness parameter 
three more times where the first two are
separated by 10s but the last parameter change 
(at 270\textsuperscript{th} second) happens after a 90s gap. 
Throughout the 5-minute video, \approach adjusts the camera setting 8 times. 
The camera setting adaption improves the mIoU per frame by 0.026 
on average with the
maximum mIoU improvement of 0.67 in comparison with using the default camera
parameter setting.
(2) \approach improves the mIoU for 24.8\% of
the video frames 
(by a maximum of 0.67)
and only minimally reduces the mIoU for 1.6\% of the frames 
(by a maximum of 0.005).
An mIoU value of zero implies that no face in the input
scene is detected by the face-detection AU. 
(3) ~\figref{fig:mIoU} also
shows that while faces are not detected under the default setting for
2.4\% of the frames, the face-detection AU can detect faces in
those frames once \approach adapts the camera parameters.

\input howquickly

\subsection{Real-world Deployment (Parking Lot)}

\begin{figure}[t]
    \centering
    \includegraphics[width=0.55\linewidth]{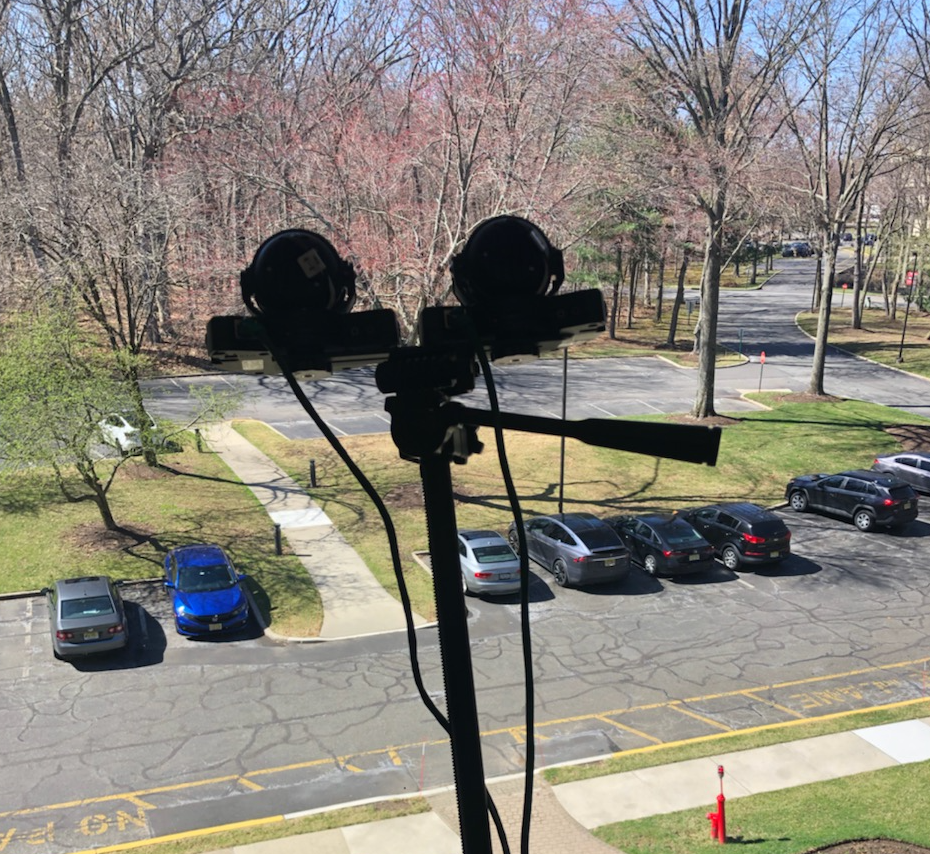}
    \vspace{-0.1in}
    \caption{\approach real-world deployment setup.}
    \label{fig:camtuner_real_setup}
\vspace{-0.1in}
\end{figure}

To validate that similar accuracy improvement from video-playback in
\S\ref{subsubsec:end2end} is achieved in real-world deployment where
the I-A parameters of the camera are continuously reconfigured by
\approach, we evaluated our deployment of \approach at a large enterprise
parking lot.  The real-world deployment has two co-located cameras, as
shown in~\figref{fig:camtuner_real_setup}. One camera is part of the
Baseline VAP (VAP 1) while the other camera is part of the \approach
VAP (VAP 2). Both VAP deployments use Axis Q3505 MK II Network
cameras, which upload the captured frames over 5G network to a remote
edge-server (with a Xeon processor and an NVIDIA GPU) running the
Efficientdet~\cite{tan2020efficientdet} object detection model to
detect cars and persons in the parking lot.
In VAP 2, the captured frames are also sent in
parallel to \approach which runs on a low-end Intel-NUC box (with a
2.6 GHz Intel i7-6770HQ CPU). \approach is seeded with the same initially VC-trained
RL agent as in \S\ref{subsubsec:end2end} and it performs a few initial online exploration rounds
and then starts exploitation and adjusts camera settings every 30 seconds. 
To evaluate the accuracy of the AUs in the VAPs, we ensure that both
cameras view almost identical scenes at the same time. 

\begin{figure}
\def\tabularxcolumn#1{m{#1}}
\begin{tabularx}{\linewidth}{@{}cXX@{}}
\begin{tabular}{cc}
\subfloat[Morning-time video]{\includegraphics[width=3.9cm]{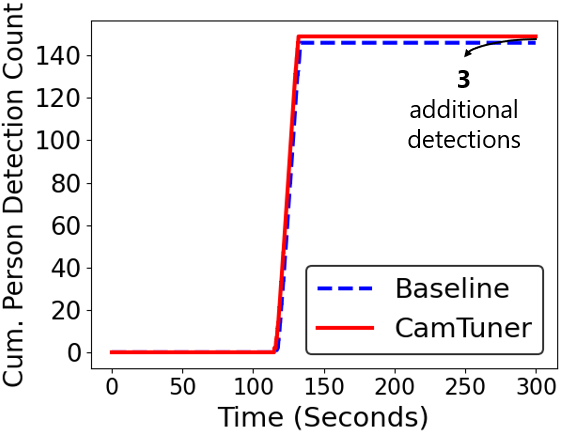}\label{fig:day_person}}
& \subfloat[Morning-time video]{\includegraphics[width=4cm]{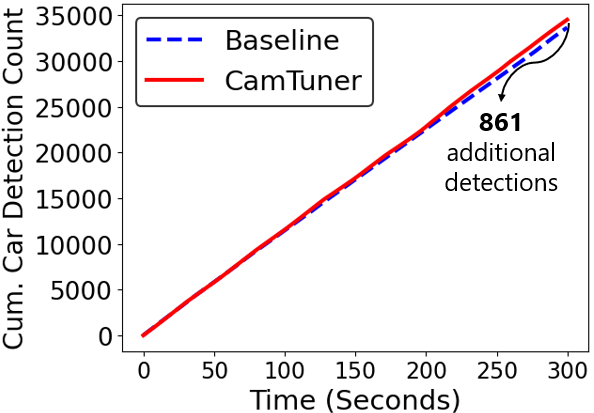}\label{fig:day_car}}\\
\subfloat[Evening-time video]{\includegraphics[width=3.98cm]{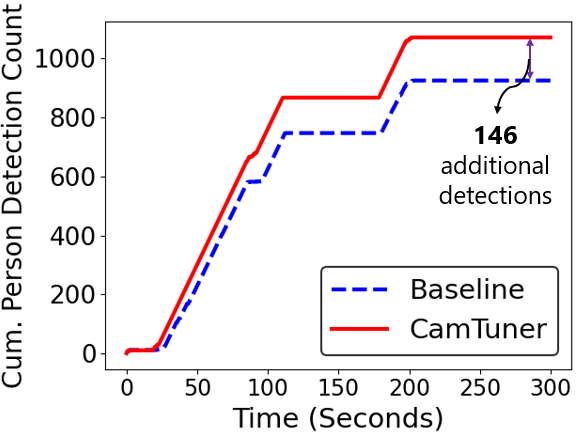}\label{fig:evening_person}}
& \subfloat[Evening-time video]{\includegraphics[width=4cm]{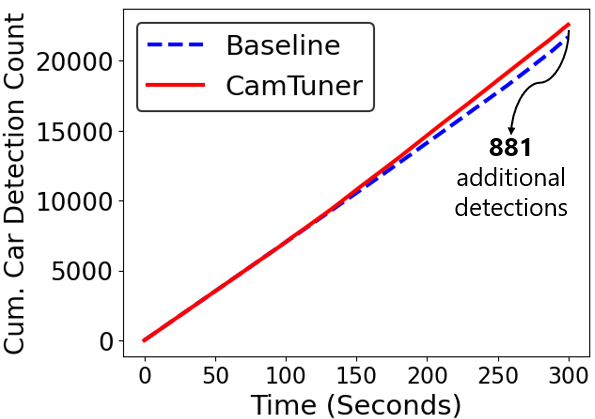}\label{fig:evening_car}} 
\end{tabular}
\end{tabularx}
    \vspace{-0.1in}
\caption{\approach performance in Parking lot.}
\vspace{-0.2in}
\label{fig:camtuner_real_eval}
\end{figure}

We ran both VAPs side-by-side for 8 continuous hours in a day and
recorded the videos from both VAPs. Since we want to manually inspect
and validate the detections from both VAPs, we randomly picked
detections for 5-minute spans during Morning and Evening time and
compare car and person detections across the two VAPs.
~\figref{fig:camtuner_real_eval} shows the cumulative number of
true-positive car and person detections. ~\figref{fig:day_person} and
~\figref{fig:evening_person} show that \approach detects 2.2\% (3) and 15.9\% (146)
additional persons than Baseline during Morning and Evening, respectively. \approach
also detects 2.6\% (861) and 4.2\% (881) more cars than the Baseline VAP during Morning and Evening,
respectively, as shown in~\figref{fig:day_car} and
~\figref{fig:evening_car}. Upon manual inspection of the videos, we
confirmed that \approach does not have any false positive detections for
car/person.

\vspace{-0.05in}
\input 5gusecase

\vspace{-0.05in}
\subsection{System Performance}
\label{subsec:overhead}

Since \approach runs in parallel with the AU,
it does not add any additional latency to the VAP and hence 
the AU latency.
In the following, we show that the normal online operation of \approach 
is light-weight,
and  the initial training phase using VC can explore each action extremely fast.

First, during online operation, each iteration of \approach involves three tasks:
evaluating the AU-specific quality estimator, 
evaluating the Q-function by the SARSA agent, 
and changing the parameters of the physical camera.
We run \approach on a low-end edge device, 
an Intel-NUC box equipped with a 2.6 GHz Intel i7-6770HQ CPU.
%
The AU-specific quality estimator takes 40ms
and the SARSA RL agent takes less than 1 ms to 
complete Q-function calculation and Q-table update.
Since the two tasks can be pipelined with changing
the physical camera settings which takes up to 200 ms on the {AXIS Q3505 MK II Network} camera
we used, each iteration of \approach takes 200 ms,
\ie 5 iterations per second, and the average CPU utilization is only 15\%
{with 150 MB memory footprint}.

Next, we run the initial RL training phase on a high-end PC with
a 3.70 GHz Intel(R) Xeon(R) W-2145 CPU and GeForce RTX 2080 GPU. 
During the one-hour training phase performed in \S\ref{subsec:VCeval},
in each iteration of the RL exploration, VC takes 4 ms to output $f_o$, 
the quality estimator takes 10 ms, 
and the RL agent take less than 1 ms to evaluate the Q-function and update the Q-table,
for a total of 15 ms.
As a result, \approach can explore around 70 actions per second,
which is 14X faster than using the physical camera.
The CPU utilization in this case is steady at 60\%.

\vspace{-0.05in}
\subsection{Accuracy of Offline Trained Models}
\label{subsec:components}

Finally, we evaluate the efficacy of two key components of \approach
which are trained offline: VC and AU-specific analytics
quality estimator model.

\begin{table}[tp]
 \caption{Accuracy of VC.}
    \vspace{-0.1in}
 \label{tab:eval-vc}
{\small
 \centering
 \begin{tabular}{|c|c|c|c|c|c|}
 \hline Parameter & Brightness & Contrast & Color & Sharpness \\ 
 & &  -Saturation & & 
 \\ \hline Mean error & \SI{5.4}{\percent} &
 \SI{13.8}{\percent} & \SI{17.3}{\percent} & \SI{19.8}{\percent}
 \\ \hline
 Std. dev. & \SI{ 1.7}{\percent} & \SI{ 4.3}{\percent} & \SI{
   9.6}{\percent} & \SI{ 8.1}{\percent} \\ \hline
 \end{tabular}
}
\vspace{-0.1in}
 \end{table}

{\bf Virtual camera.}
\label{subsec:VCeval}
VC is designed to render a frame taken at one time ($T_1$) to 
another time ($T_2$), as if the rendered frame were captured at time
$T_2$. First, we trained VC in the offline profiling phase as
discussed in \S\ref{subsec:VC} using a 24-hour long video obtained from one of our customer locations at an airport.
To evaluate how well VC works online, we obtained several video
snippets at {6} different hours of the day from the same camera. %
Next, we fed 1 video snippet $VS_0$ from one particular hour $H_0$
through VC which applies different digital transformation to
generate 5 video snippets $VS_{j}$ corresponding to the hours of the
other 5 videos. For each generated video snippet $VS_j$, we calculated
the relative error of the metric tuple values of each frame in
$VS_{j}$ relative to that of the corresponding original video frame
and average such error across all the frames in $VS_j$ {(over 37.5K
  frames)}. We obtained 5 VC error metric tuples for one video, each
corresponding to the hour of the other 5 video snippets.  We repeated
the above experiment for the 5 other original video snippets to obtain
a total of 30 VC error tuples.
~\tabref{tab:eval-vc} shows the mean error and standard
deviation among {all 30 VC error tuples. 
We observe that the average VC errors 
are \emph{5.4\%}, \emph{13.8\%}, \emph{17.3\%}, and
\emph{9.8\%} for brightness, contrast, color-saturation and sharpness,
respectively.}
 
{\bf AU-specific analytics quality estimator.}
\begin{figure}[t]
    \centering
     \includegraphics[width=0.65\linewidth]{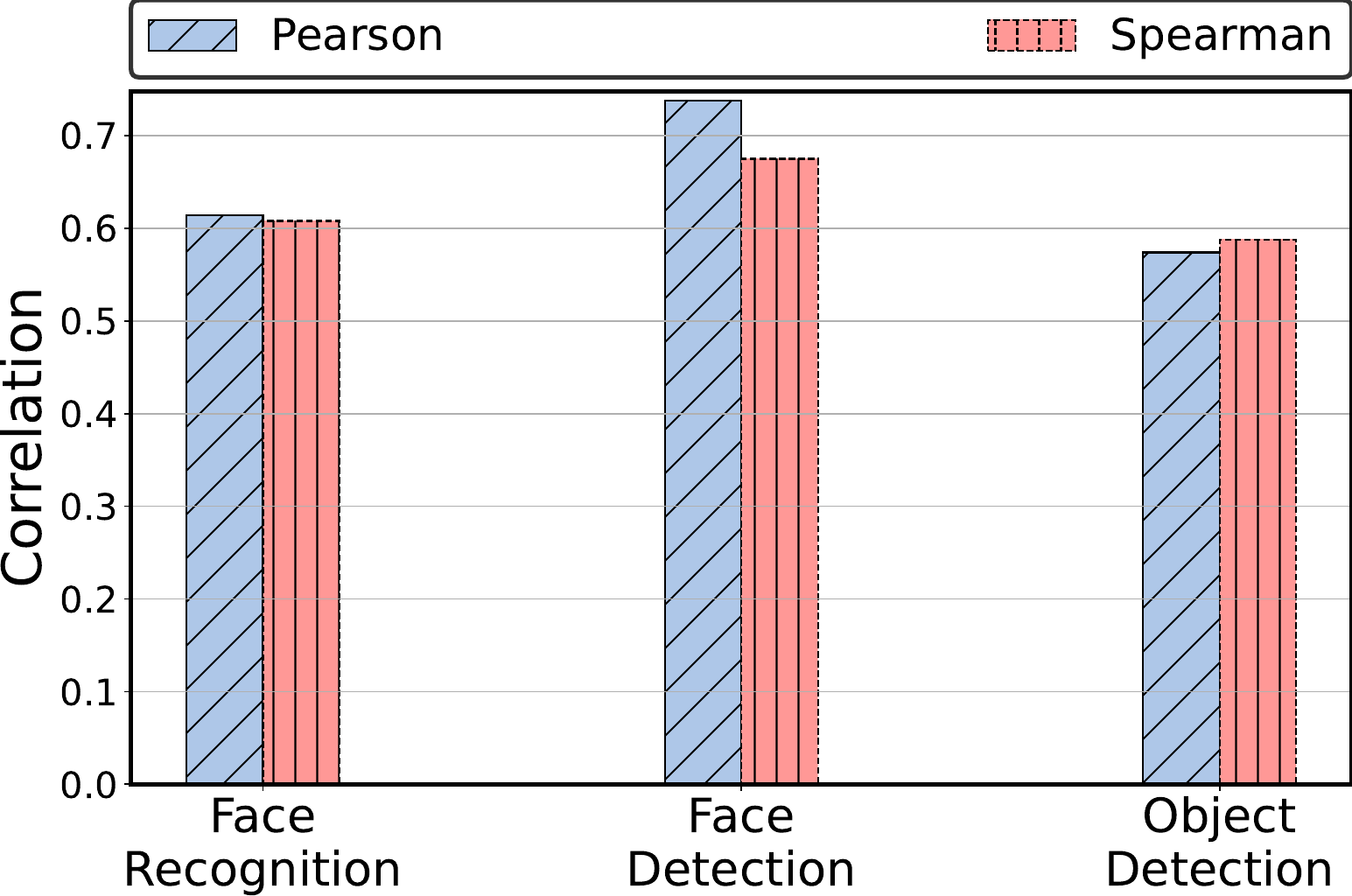}
    \vspace{-0.1in}
    \caption{Analytics quality estimator performance.}
    \label{fig:au_corr}
     \vspace{-0.25in}
\end{figure}
%
Next, we evaluate the performance of AU-specific quality estimators.
Since the AU-specific estimator is a lightweight model that
  predicts coarse-grained accuracy measure of the heavyweight DNN
  model (\ie used in AU), it is not meaningful to compare its
  accuracy against the accuracy achieved by the heavyweight
  model (derived using ground truth).
Instead, we measure the quality of 
the AU-specific quality estimator by 
measuring the Spearman and Pearson correlation between 
the two accuracies
for three different AUs i.e. face-recognition, face-detection, and
  person-detection.
%
First, we trained the three estimators through supervised learning as described
in~\secref{sec:AU-specific}.
To evaluate the face-recognition estimator,
we used the celebA-validation dataset which contains 200 images (\ie\ different from the 300 original training images used in~\secref{sec:AU-specific}) and their about 2 million variants 
{from augmenting the original images using the python-pil image library~\cite{pil}}.
~\figref{fig:au_corr} shows that
the quality predicted by the face-recognition analytics quality
estimator is strongly correlated with the output by the AU 
(both Pearson and Spearman correlation are greater than 0.6)~\cite{corr, corr_1}. 

To evaluate the face-detection quality estimator, we used annotated video
frames from the \textit{olympics}~\cite{niebles2010modeling} and
\textit{HMDB} datasets~\cite{Kuehne11} and their 4 million variants
that were generated.
To evaluate object-detection analytics quality estimator,
we used labelled images (\ie only consist car and
person object classes) from the COCO dataset~\cite{lin2014microsoft}
and their 7 million augmented variants 
~\figref{fig:au_corr} shows that there is a strong
positive correlation between the measured mAP and IoU metric and the
predicted quality estimate for both face-detection and
object-detection AUs. 
In summary, the strong correlation between the prediction by the
estimators and the actual quality of AUs based on ground truth,
enables \approach's RL agent to effectively tune camera parameters.

%% file: howquickly.tex
\subsection{How quickly does \approach react to suboptimal settings}

\begin{figure}
\begin{subfigure}[t]{0.3\linewidth}
\centering
    \includegraphics[width=\linewidth]{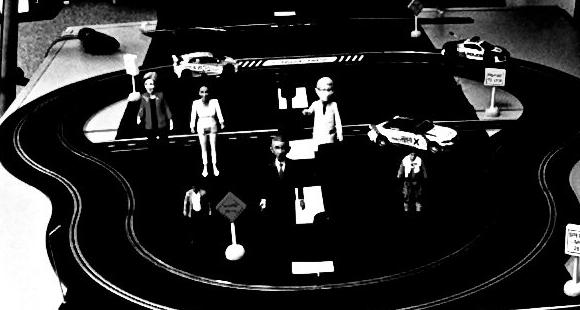}
    \caption{capture under SS1}
    \label{fig:static-stream-1}
\end{subfigure}
\begin{subfigure}[t]{0.3\linewidth}
\centering
    \includegraphics[width=\linewidth]{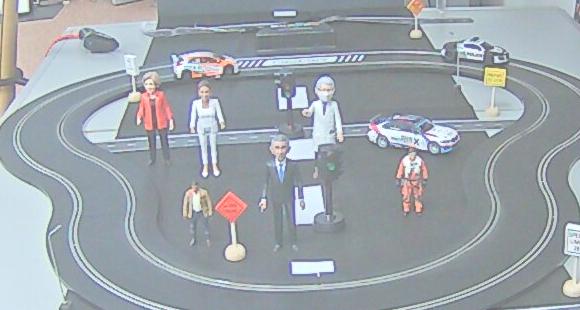}
    \caption{capture under SS2}
    \label{fig:static-stream-2}
\end{subfigure}
\begin{subfigure}[t]{0.3\linewidth}
\centering
    \includegraphics[width=\linewidth]{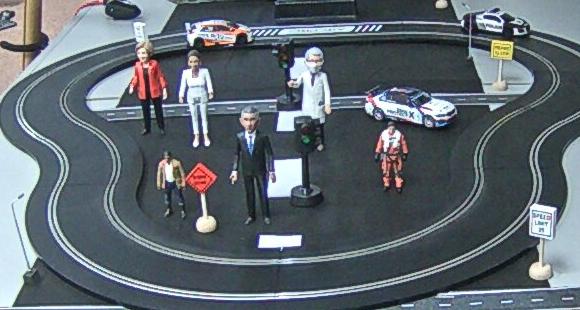}
    \caption{\approach camera capture}
    \label{fig:camtuner-stream}
\end{subfigure}%
    \vspace{-0.1in}
 \caption{Sample static \& \approach camera captures.}
  \label{fig:camtuner-static-stream}
 \vspace{-0.1in}
\end{figure}

\begin{figure}
    \centering
    \begin{subfigure}[t]{0.495\columnwidth}
        \vskip 0pt
        \centering
        \includegraphics[width=0.99\textwidth]{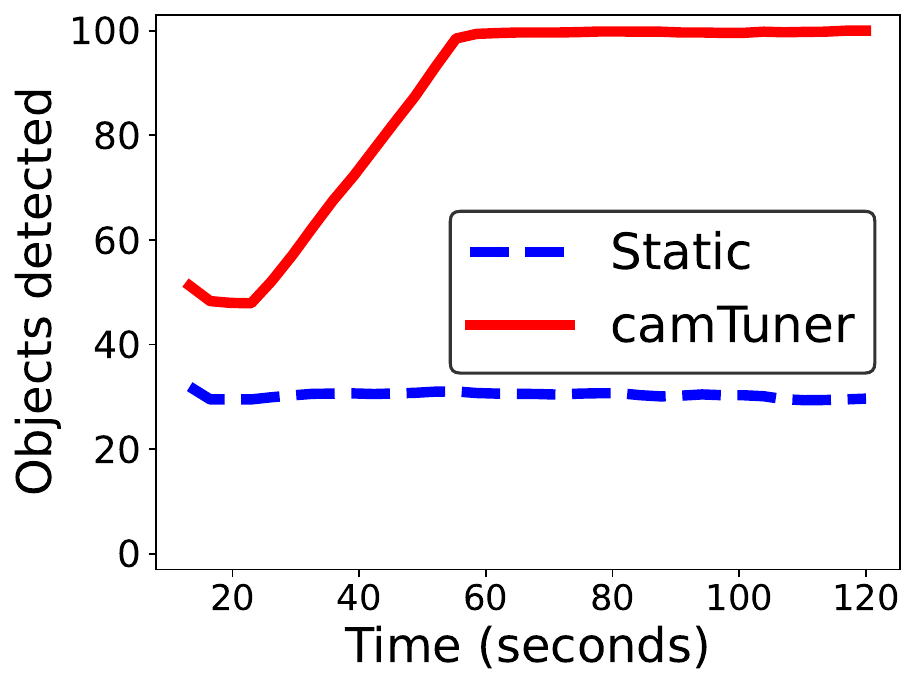}
        \caption{suboptimal setting 1 (SS1)}
        \label{fig:suboptimal_1}
    \end{subfigure}
    \hfill
    \begin{subfigure}[t]{0.495\columnwidth}
        \vskip 0pt
        \centering
        \includegraphics[width=0.99\textwidth]{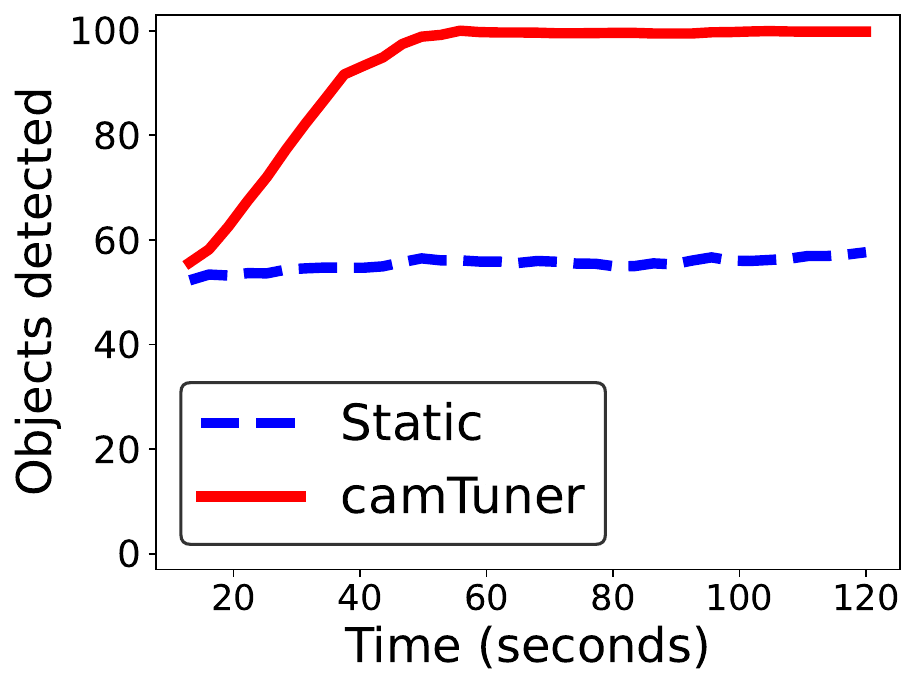}
        \caption{suboptimal setting 2 (SS2)}
        \label{fig:suboptimal_2}
    \end{subfigure}
    \vspace{-0.1in}
\caption{\approach reaction to suboptimal settings (Normalized Moving average of total object detection is computed over last 100 frames, shown in Y axis.)}
\label{fig:eval_suboptimal}
\vspace{-0.2in}
\end{figure}

Here, we evaluate how quickly \approach can react if the
  camera is set to a suboptimal setting that leads to degraded
  analytical outcome. We place two side-by-side cameras in
  front of a scene consisting of 3D objects as shown
  in~\figref{fig:camtuner-static-stream}. In this scene, 3D slot cars
  are continuously moving over the track and 3D human models are kept
  stationary.
Both cameras start with a same suboptimal setting (we use two suboptimal settings denoted as SS1 and SS2) and stream at 10 FPS for 2-minute period,
  during which the I-A parameters of Camera 1 are kept to the same initial
  suboptimal values, while the I-A parameters of Camera 2 are tuned by
  \approach every 2s.
{On every frame streamed from camera}, we use Yolov5~\cite{glenn_jocher_2022_6222936} as the object detector to detect and
  record the type of objects with their bounding boxes~\footnote{Manual
    inspection confirms no false-positive detection in the 2-minute
    period.}.
~\figref{fig:eval_suboptimal} plots the normalized moving average of the total number of object detections in the last 100 frames in the Y axis (to clearly show
the trend) of the two cameras under two different initial suboptimal
settings, SS1 and SS2.
We observe a small initial gap between the performance of YOLOv5 between the two camera streams
which indicates that within the first 10 seconds,
\approach changes the camera
  parameters once based on analytics quality estimator output and 
  achieves better object detection.
Furthermore, we observe that
\approach gradually converges to a best-possible setting
within a minute that enables Yolov5 to detect all
objects from the scene (total 5-7 more object detections per frame).

%% file: 5gusecase.tex
\subsection{5G Use Case: Automatic Vehicle Collision Prediction (AVCP)}
\label{subsec:accipreven}
\begin{figure}
\begin{subfigure}[t]{0.282\linewidth}
\centering
    \includegraphics[width=\linewidth]{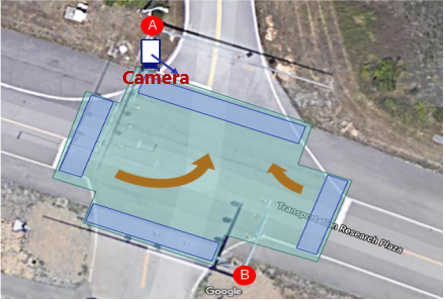}
    \caption{Intersection}
    \label{fig:intersec}
\end{subfigure}%
\begin{subfigure}[t]{0.34\linewidth}
\centering
    \includegraphics[width=\linewidth]{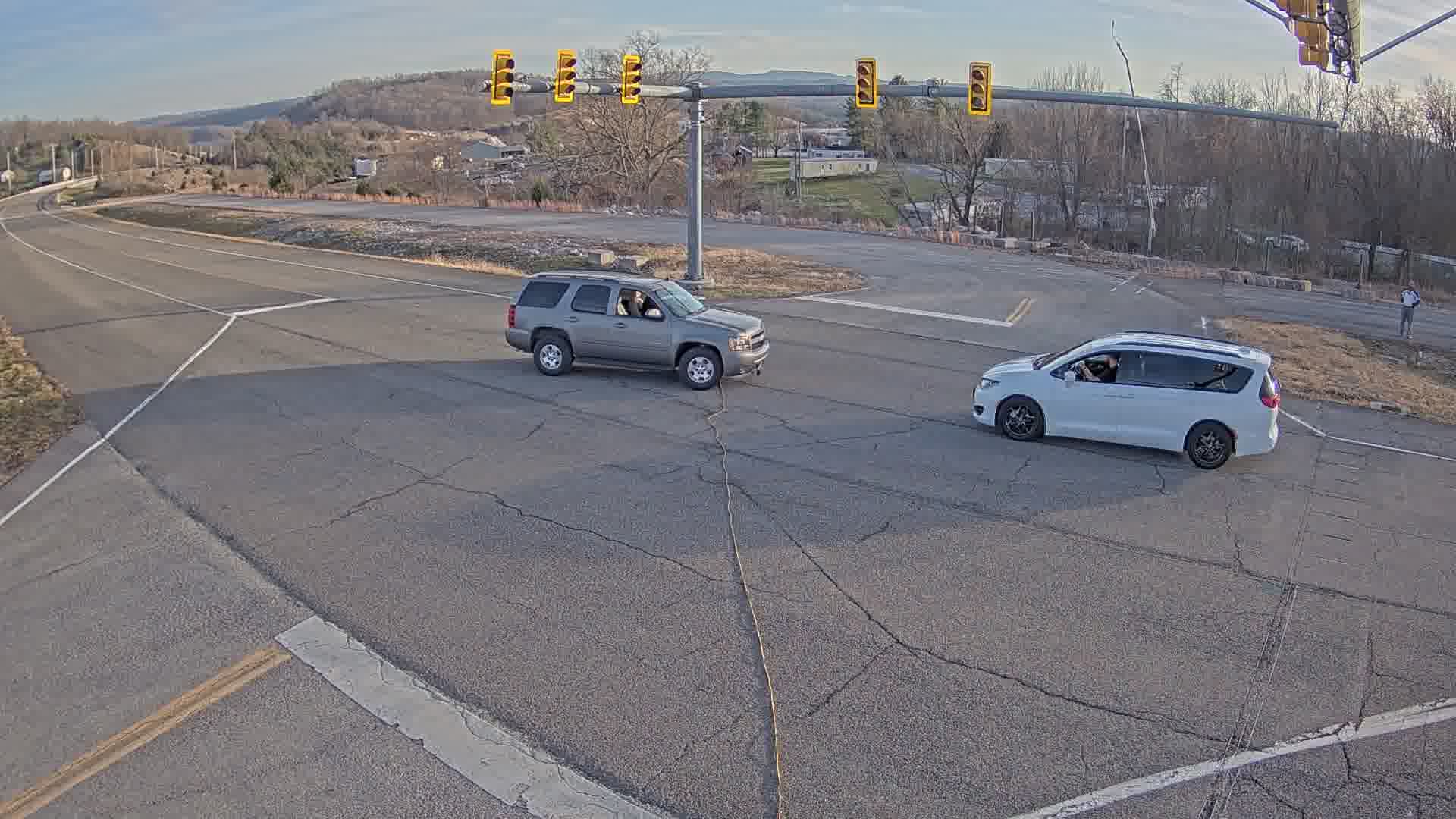}
    \caption{Before accident}
    \label{fig:before_acc}
\end{subfigure}
\begin{subfigure}[t]{0.34\linewidth}
\centering
    \includegraphics[width=\linewidth]{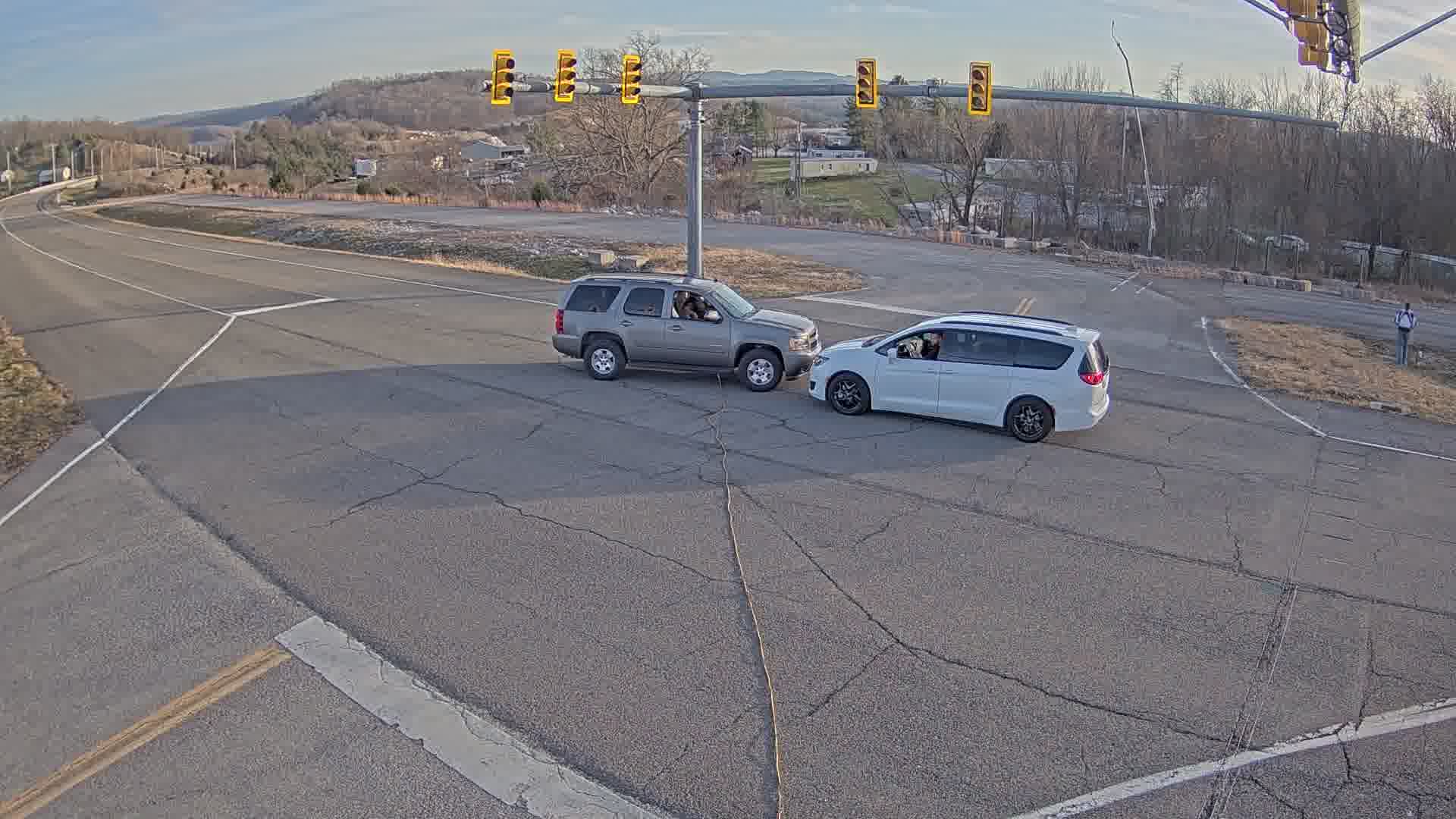}
    \caption{Car accident}
    \label{fig:after_acc}
\end{subfigure}
    \vspace{-0.1in}
 \caption{Car accident prevention scenario.}
  \label{fig:car-accident-scenario}
 \vspace{-0.2in}
\end{figure}

\begin{figure}[t]
    \centering
    \includegraphics[width=0.75\linewidth]{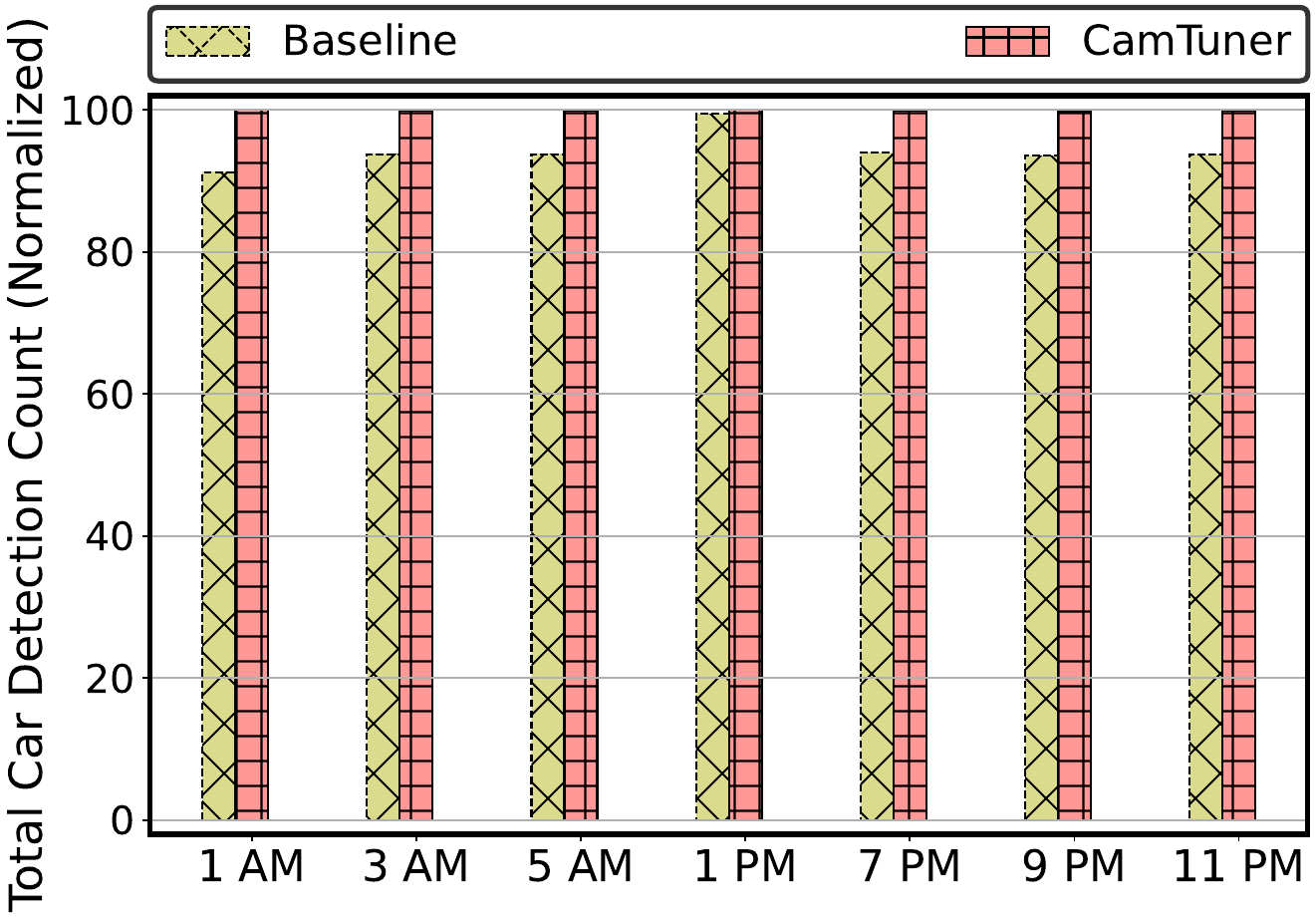}
    \caption{\approach performance in \textit{Accident Prevention} scenario.}
    \label{fig:accident_camtuner}
\vspace{-0.2in}
\end{figure}

AVCP is an important use case in Intelligent
Transportation Systems (ITS). This use case requires extremely low
latency because it is very critical to predict collision and react almost
instantaneously in order to prevent a potential life-threatening accident. 
In particular, low latency in the
order of milliseconds is desired for this use case which can be
achieved by using 5G, which promises ultra-reliable low latency
communication (URLLC). 
In AVCP usecase, reliably detecting and tracking vehicles and pedestrians is
one of the most critical building blocks; without this, the collision
prediction won't work properly and may lead to life-threatening accidents.
\approach\ plays an important role in AVCP usecase, where it changes the
camera settings dynamically in reaction to the environmental
changes. Since the environment does not change within seconds,
\approach's quick (in the order of milliseconds) adjustment of camera
setting improves the detection and tracking of vehicles and
pedestrians at all times.

To evaluate this use case, we recorded a 1.5-minute-long car accident
scenario, as shown in ~\figref{fig:car-accident-scenario}, at one of
our customer sites that has a 5G smart traffic intersection
testbed. We then used the experimental methodology described in
\S\ref{subsubsec:expsetup} to emulate the same car accident scenario
at 7 different hours (\ie environmental conditions) of the same day,
and running 2 VAPs (VAP 1 uses the default camera settings, denoted as Baseline and VAP 2 runs \approach) that detect cars using \textit{EfficientDet} object
detector.  ~\figref{fig:accident_camtuner} shows that compared to VAP
1, VAP 2 using \approach detects
6.2\% additional cars on average across the different hours,
and as much as 9.7\% (122) more cars at 1 AM.

\cut {
To observe the impact of camera parameter tuning on true-positive car
detection count and ultimately possible accident prediction, we collected
several videos that capture real-world car driving patterns and
possible car crashing scenarios on major streets from a camera
located at a traffic intersection, as shown
in~\figref{fig:car-accident-scenario}. Here, the drivers are instructed to
drive the vehicles accordingly that emulate the real-world car
accident scenarios. Figures\ref{fig:before_acc}--\figref{fig:after_acc} show how the camera captures the positions of
two cars right before and after the accident happened, respectively. The
accident prevention AU detects the car bounding boxes across
the frames and monitors their trajectory; if the two cars' bounding boxes
are sufficiently closeby, it raises an accident alert.

{Due to data collection difficulty on major streets and not
  having ground-truths under real-world deployment}, we use the same
experimental methodology as described
in~\secref{subsubsec:expsetup}. We use the \textit{EfficientDet}
object detector to detect cars from the input scene.  For different
environmental effects augmented by the VC, dynamically tuning the
camera setting by \approach reduces the total true-positive car
detection count by 2.1\% (29 fewer detections) at 1pm and 8.85\% (122
fewer detections) at 1am, as shown by the green bars
in~\figref{fig:accident_camtuner}.

The above result shows that $\approach$ can adapt the
camera parameters for different environmental conditions and improve
the input video quality where the AU can accurately detect the cars across
the frames. By detecting cars more often, \approach helps to
more accurately monitor the trajectory of the cars across the frames which in turn
results in accurate prediction of the accident occurrence in advance~\cite{???}.
}


\cut{
\begin{figure*}
\begin{subfigure}[t]{0.164\textwidth}
\centering
    \includegraphics[width=\textwidth]{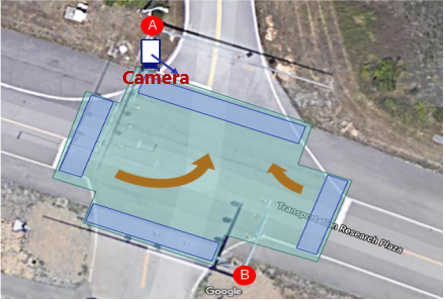}
    \caption{Camera at traffic intersection}
    \label{fig:intersec}
\end{subfigure}%
\begin{subfigure}[t]{0.2\linewidth}
\centering
    \includegraphics[width=\textwidth]{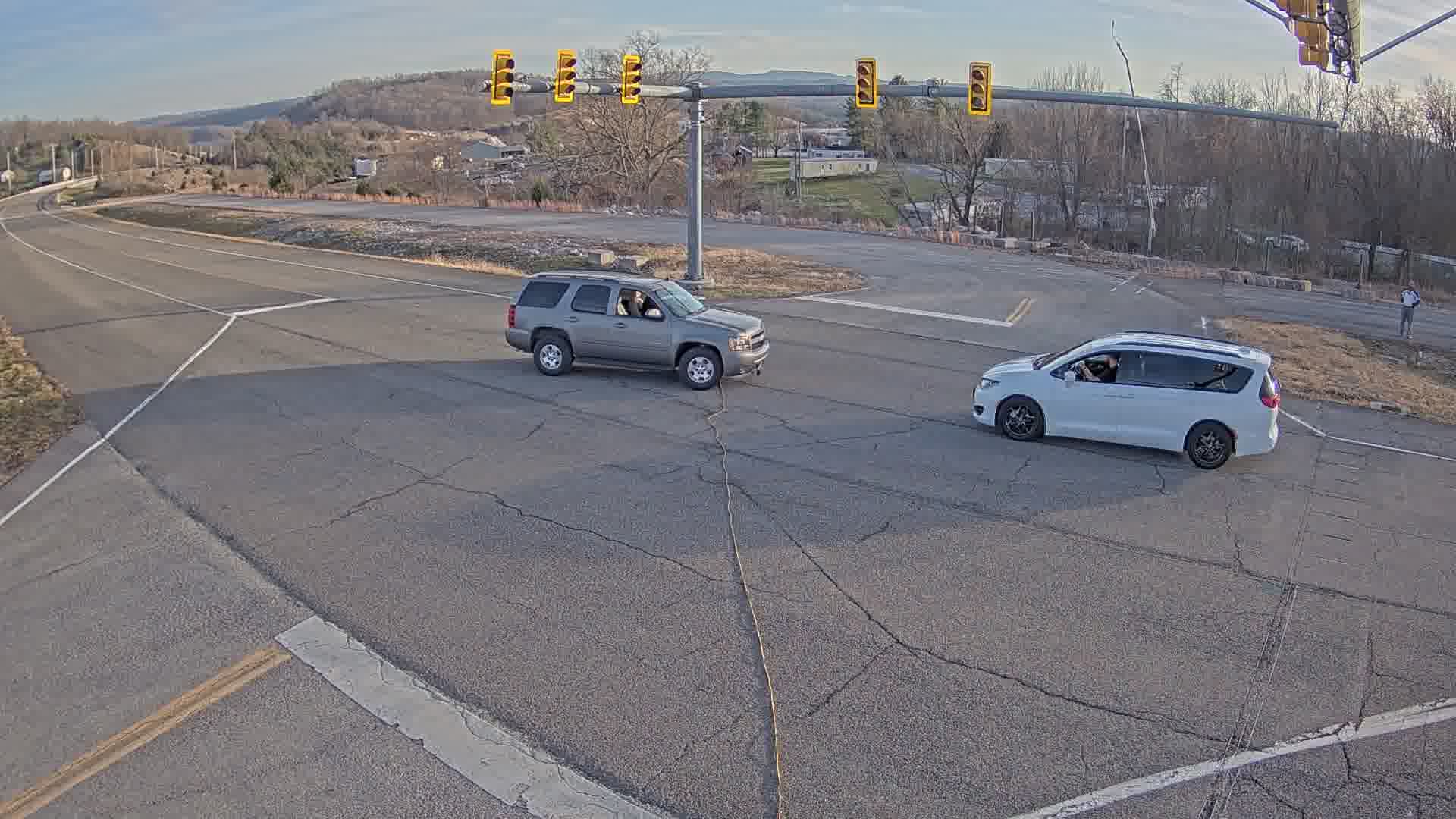}
    \caption{Before accident}
    \label{fig:before_acc}
\end{subfigure}
\begin{subfigure}[t]{0.2\linewidth}
\centering
    \includegraphics[width=\textwidth]{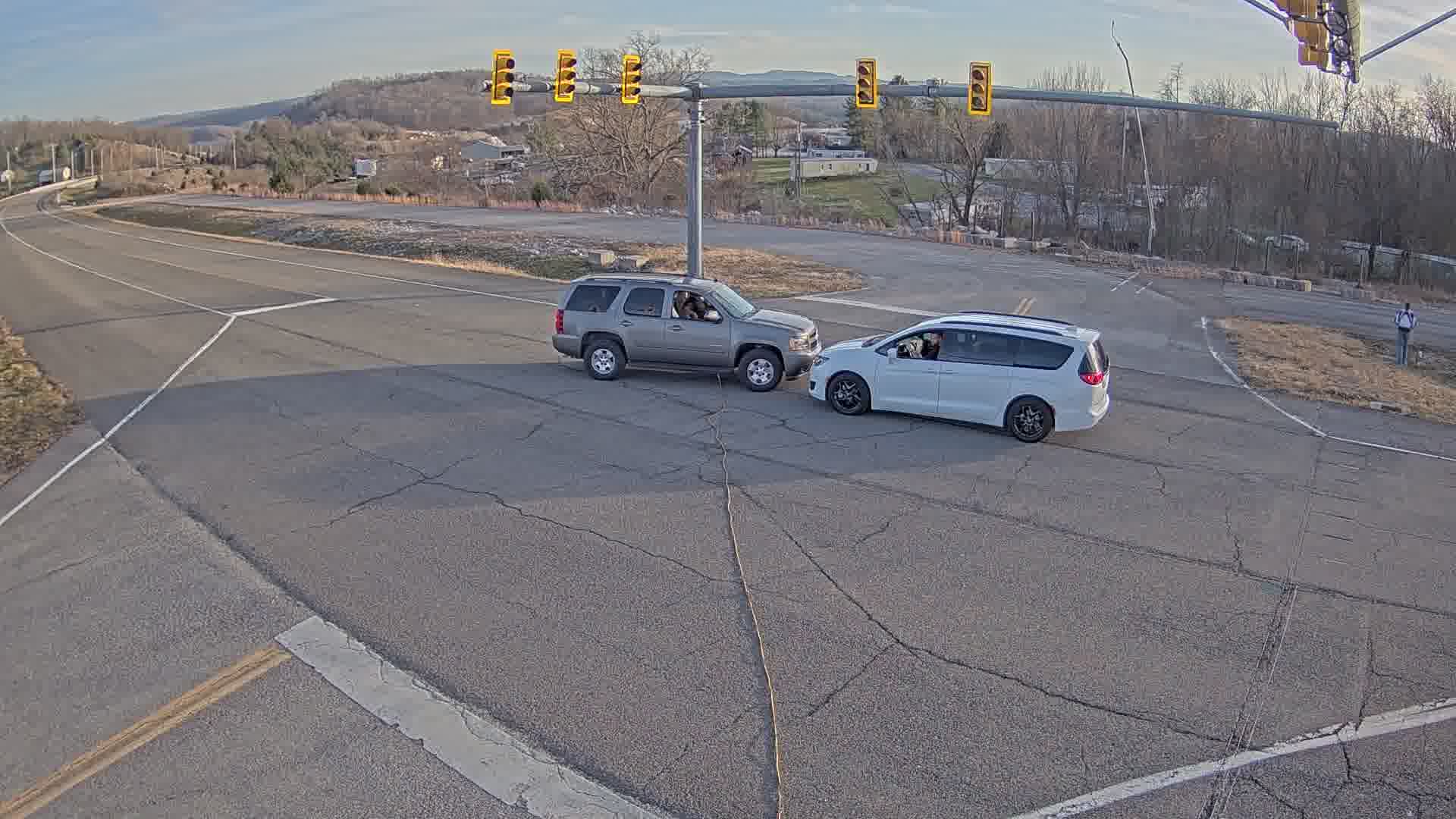}
    \caption{Accident happened}
    \label{fig:after_acc}
\end{subfigure}
 \caption{Car accident prevention scenario.}
  \label{fig:car-accident-scenario}
\end{figure*}
}



%% file: related.tex
\section{Related Work}
\label{section:related}


Similar to our findings, Jang \etal \cite{Jang2018ApplicationAwareIC}
also show that environmental condition changes affect VAP, but their
approach to adapt to such changes is to use different AUs depending on
the environmental conditions, \eg using Haar cascade for detection when
lighting is sufficient and switch to HOG when environment gets
darker. Since there could be several reasons for change in environment
as discussed in \S\ref{subsec:environ}, developing an AU for every kind
of environment
is not feasible. In contrast, \approach takes a different approach
where the AU is fixed and camera settings are adjusted to adapt to
environmental changes. 

Several works investigate tuning parameters of a VAP after camera
capture and before sending it to an AU or changing the AU based on the
input video content.  Videostorm~\cite{201465},
Chameleon~\cite{jiang2018chameleon}, and Awstream~\cite{zhang2018awstream}
tune the after-capture video stream parameters such as frames-per-second or
frame resolution to ensure efficient resource usage while
processing video analytics queries at scale.
In contrast, \approach dynamically tunes camera parameters to improve
the AU accuracy of VAPs.

More recent work, \eg Focus~\cite{222587},
NoScope~\cite{10.14778/3137628.3137664},
Ekya~\cite{padmanabhantowards}, and AMS~\cite{khani2020real}, studied
how to adapt AU model parameters based on captured video content.
Such an approach requires additional GPU resources for periodic
model retraining
and is also less reactive to video
content changes.  In contrast, \approach quickly adapts the camera
parameters in real-time according to environmental changes.

Several frame filtering techniques on edge
devices~\cite{MLSYS2019_85d8ce59, paul2021aqua, chen2015glimpse,
li2020reducto} can work in conjunction with \approach\ and potentially further
improve \approach's performance. Our AU-specific analytics quality
estimator shares similar goal as the AQuA-quality
estimator~\cite{paul2021aqua} but differs in that
\approach's quality estimator performs quality estimation that is specific to each AU, while AQuA performs much more coarse-grained AU-agnostic image quality estimation.

There is a large body of work on configuring the Image Signal
Processing pipeline (ISP) in cameras to improve human-perceived
quality of images from the cameras 
~\cite{wu2019visionisp, liu2020joint, diamond2021dirty, nishimura2019automatic}.
In contrast, we study dynamic camera parameter tuning to optimize the accuracy of VAPs.

\cut{
(2) the two also differ in training labels
generation for respective quality estimators. 
In the case of \emph{AQuA}, training labels are generated from passing the input image through a bank of classifiers; for \approach, training labels are dependent on how each AU functions and their preferred way of measuring accuracy.
}

%% file: future.tex
\section{Future Work}
\label{section:future}


%
%

Our first demonstration in this paper of how NAUTO camera parameters
can be dynamically tuned to enhance video analytics accuracy opens up
many avenues for future research. Here, we consider only a subset of
NAUTO camera parameters related to image appearance but there are
several other NAUTO parameters such as max exposure time, maximum gain, and
defog effect (listed in ~\tabref{tab:cameras}) that can also be
dynamically tuned to enhance video analytics accuracy. In our future
work, we plan to study the impact of dynamically tuning these other
camera parameters as well.

Current \approach design tunes NAUTO camera parameters in order to
enhance video analytics accuracy of a single AU. However, in a typical
camera deployment there could be multiple AUs sharing a single video
stream. For example, in an enterprise building entrance where a
surveillance camera is deployed, it is common to have separate AUs
including authorization for the car and people entering, counting number
of cars and people entering/leaving, \etc 
that all need to run on the same camera stream. In our future work, we
will study how to tune camera parameters in order to simultaneously
enhance the accuracy for multiple AUs.




%% file: conclusion.tex
\section{Conclusion}
\label{section:conclusion}

In this paper, we presented the design and evaluation of \approach, 
to our knowledge the first
adaptive VAP framework that 
adaptively learns the best setting 
for its NAUTO camera parameters deployed in the field in reaction to
environmental changes to enhance AU accuracy.
Our controlled experiments and real-world VAP deployment
show that compared to a VAP using the default camera setting,
\approach allows the VAP to detect 
15.9\% additional persons 
and 2.6\%--4.2\% additional cars 
(without any false positives)
in a large enterprise large parking lot
and 9.7\% additional cars in a 5G smart traffic intersection scenario.
\approach dynamically 
determines how to tune IP camera parameters, which
  can be executed either directly inside the camera via the exposed
  REST APIs for remotely configuring the camera setting, or via digital
  transformation after camera capture, \eg for cameras that do not
  expose such APIs.
Furthermore, we believe \approach's design and its key components,
\emph{Virtual Camera} and \emph{light-weight AU-specific analytics
  quality estimators}, can be applied to dynamically tune other
complex sensors such as depth and thermal cameras.

\cut{
In this paper, we showed that environmental condition changes can adversely affect accuracy of video analytics pipelines and this accuracy loss can be mitigated by changing camera settings. Since it is not practical for humans to constantly monitor and change these camera settings in real-time, we proposed \approach, which dynamically adapts to the changes in environmental conditions by automatically adjusting camera settings in real-time and improves AU accuracy. We show that in a real-world deployment with two side-by-side cameras, \approach detects $\sim$ 875 additional cars and $\sim$ 150 additional persons (across frames within a 5 minute span). Also, for a real-world accident prevention 5G usecase, using a recorded accident scenario scene, we show that \approach detects $\sim$ 125 additional cars (across frames within a 1.5 minute span), which directly aids in preventing a potential catastrophic car accident.
}

\cut{
In this paper, we showed that in a typical surveillance camera
deployment, environmental condition changes can significantly affect
the accuracy of analytics units in a VAP.  We developed \approach, a
system that dynamically adapts complex settings of the camera in a VAP
to changing environmental conditions to improve the AU accuracy.  
Through dynamic tuning, \approach is able to achieve up to 13.8\% and 9.2\% higher accuracy for the face detection AU and person detection AU respectively, compared to the best of the two approaches i.e., baseline and strawman approach (average improvement of $\sim$8\% for both face detection AU and person detection AU).
The \approach-enhanced VAP has been running at our customer sites since Summer 2021, and shown to improve the accuracy of the face detection AU by 11.7\% compared to the original VAP without \approach. We believe \approach's system design and key components, Virtual camera and light-weight AU-specific analytics quality estimators, can be applied to dynamically tune other complex sensors such as depth and thermal cameras.
}